\pdfoutput=1

\documentclass[11pt]{article}

\usepackage{ACL2023}

\usepackage{times}
\usepackage{latexsym}

\usepackage[T1]{fontenc}

\usepackage[utf8]{inputenc}
\usepackage[T2A,LAE,T1]{fontenc}
\usepackage[arabic,USenglish]{babel}
\usepackage{CJKutf8}

\usepackage{microtype}

\usepackage{inconsolata}

\usepackage{booktabs}
\usepackage{multirow}
\usepackage{subcaption}
\usepackage{changepage}
\usepackage[dvipsnames]{xcolor}

\definecolor{p140_green}{RGB}{0,150,0}
\definecolor{p140_red}{RGB}{250,128,114}
\definecolor{pred_dark_orange}{RGB}{230,97,1}
\definecolor{pred_light_orange}{RGB}{253,184,99}
\definecolor{pred_purple}{RGB}{94,60,153}
\definecolor{pred_blue_bell}{RGB}{178,171,210}

\usepackage{graphicx}
\usepackage{listings}
\usepackage{rotating}

\usepackage{amssymb}
\usepackage{bm}
\usepackage{float}
\title{DLAMA: A Framework for Curating Culturally Diverse Facts for Probing the Knowledge of Pretrained Language Models}

\author{Amr Keleg and Walid Magdy\\
  Institute for Language, Cognition and Computation \\
  School of Informatics, University of Edinburgh \\
  \texttt{a.keleg@sms.ed.ac.uk}, 
  \texttt{wmagdy@inf.ed.ac.uk} \\}

\begin{document}

\maketitle

\begin{abstract}
A few benchmarking datasets have been released to evaluate the factual knowledge of pretrained language models. These benchmarks (e.g., LAMA, and ParaRel) are mainly developed in English and later are translated to form new multilingual versions (e.g., mLAMA, and mParaRel).
Results on these multilingual benchmarks suggest that using English prompts to recall the facts from multilingual models usually yields significantly better and more consistent performance than using non-English prompts.
Our analysis shows that mLAMA is biased toward facts from Western countries, which might affect the fairness of probing models. We propose a new framework for curating factual triples from Wikidata that are culturally diverse.
A new benchmark \textbf{DLAMA-v1} is built of factual triples from three pairs of contrasting cultures having a total of 78,259 triples from 20 relation predicates. The three pairs comprise facts representing the (Arab and Western), (Asian and Western), and (South American and Western) countries respectively. Having a more balanced benchmark (DLAMA-v1) supports that mBERT performs better on Western facts than non-Western ones, while monolingual Arabic, English, and Korean models tend to perform better on their culturally proximate facts. Moreover, both monolingual and multilingual models tend to make a prediction that is culturally or geographically relevant to the correct label, even if the prediction is wrong.

\end{abstract}

\section{Introduction}
\label{introduction}
Transfer learning paradigms such as fine-tuning, few-shot learning, and zero-shot learning rely on pretrained language models (PLMs), that require having large compilations of raw data (\citealt{devlin-etal-2019-bert}; \citealt{brown2020language}; \citealt{51308}; \citealt{scao2022bloom}). These PLMs showed some ability to model different linguistic phenomena (\citealt{Goldberg2019AssessingBS}; \citealt{jawahar-etal-2019-bert}) in addition to memorizing facts related to real-world knowledge. While there is a drive to have multilingual models, English is still the language that is better supported due to the abundance of large English raw corpora, diverse datasets, and benchmarks. Moreover, monolingual non-English PLMs are still being pretrained for other high-resource languages. As a way to probe the non-English and multilingual PLMs, researchers tend to translate English benchmarks into other languages, which might degrade the quality of the samples especially if the translation is performed automatically. While translating English benchmarks saves the time and money needed to build new language-specific benchmarks, it might introduce unintended biases or artifacts into the benchmarks.

LAMA \cite{petroni-etal-2019-language} and ParaRel \cite{elazar-etal-2021-measuring} are two benchmarks developed to quantify the factual knowledge of the English PLMs. They used a setup in which a language model is said to know a specific fact if it can predict the right object for a prompt in a fill-the-gap setup (e.g., For the prompt \textbf{``The capital of England is [MASK]"}, the model needs to fill the masked gap with \textbf{``London"}). Multilingual versions of these benchmarks namely: mLAMA \cite{kassner-etal-2021-multilingual}, and mParaRel \cite{fierro-sogaard-2022-factual} were released to evaluate the performance of multilingual PLMs by translating LAMA and ParaRel into 53 and 46 languages respectively. The subjects and objects of the triples within these benchmarks were translated using their multilingual labels on Wikidata, while the templates were automatically translated from the English ones used in the original benchmarks. These templates transform triples into textual natural language prompts for probing the models. X-FACTR is another benchmark sharing the same setup, and is built for 23 different languages \cite{jiang-etal-2020-x}. All three benchmarks sample factual triples in the form of (subject, relation predicate, object) from T-REx, a dump of Wikidata triples aligned to abstracts extracted from the English Wikipedia \cite{elsahar-etal-2018-rex}. The way T-REx is constructed might make it more representative of the facts related to Western cultures, which might introduce an unnoticed bias to the benchmarks based on it. We hypothesize that having a fair representation of the different cultures within a benchmark is vital for fairly probing models pretrained for multiple languages. The main contributions of our paper can be summarized as follows:

\vspace{-0.2cm}
\vspace{-\topsep}
\begin{enumerate}
      \setlength{\parskip}{0pt}
      \setlength{\itemsep}{0pt}
    \item Investigating the impact of sampling mLAMA triples from T-REx on the distribution of the objects within the relation predicates.
    \item Proposing DiverseLAMA (DLAMA), a methodology for curating culturally diverse facts for probing the factual knowledge of PLMs, and building 3 sets of facts from pairs of contrasting cultures representing the (Arab-West), (Asia-West), and (South America-West) cultures, to form DLAMA-v1\footnote{The DLAMA-v1 benchmark and the codebase can be reached through: \url{https://github.com/AMR-KELEG/DLAMA }}.
    \item Showing the impact of having a less skewed benchmark DLAMA-v1 on the performance of mBERT and monolingual Arabic, English, Korean, and Spanish BERT models.
    \item Demonstrating the importance of having contrasting sets of facts in diagnosing the behavior of the PLMs for different prompts.
\end{enumerate}

\vspace{-\topsep}
\vspace{-\topsep}

\section{Related Work}
\label{background}

\citet{petroni-etal-2019-language} investigated the possibility of using PLMs as potential sources of knowledge, which can later substitute manually curated knowledge graphs. To this end, they created LAMA (LAnguage Model Analysis), a dataset of 34,000 relation triples representing facts from 41 different Wikidata relation predicates. These facts are extracted from a larger dataset called T-REx that contains 11 million relation triples, acquired from a large Wikidata dump of triples, that were automatically aligned to English Wikipedia abstracts \cite{elsahar-etal-2018-rex}. Manual English templates were written to transform the triples into prompts to probe the model's factual knowledge. The triples were limited to the ones whose objects are tokenized into a single subtoken.

\citet{kassner-etal-2021-multilingual} constructed a multilingual version of LAMA (mLAMA) having 53 different languages. They handled the limitation of using single-subtoken objects by computing the probability of a multi-subtoken object as the geometric mean of the subtokens' probabilities. They concluded that the performance of mBERT when probed with prompts written in 32 languages is significantly lower than mBERT's performance when probed with English prompts. Moreover, they observed insignificant performance improvement for German, Hindi, and Japanese when their corresponding templates were manually corrected. 

Similarly, \citet{jiang-etal-2020-x} created X-FACTR by sampling relation triples from T-REx for 46 different Wikidata predicates. The multilingual Wikidata labels were used to translate the subjects and objects of the triples. They compared multiple decoding methods. Moreover, they employed different templates to generate prompts having the correct number/gender agreement with the subjects of the triples. English prompts still outperformed prompts written in 22 other languages.

ParaRel and its multilingual version mParaRel are benchmarks created by sampling triples from T-REx for 38 relation predicates
(\citealt{elazar-etal-2021-measuring}; \citealt{fierro-sogaard-2022-factual}).
Their aim is to measure the consistency of the model in making the same prediction for different paraphrases of the same template. Results on both benchmarks showed that the multilingual mBERT and XLM-R models are less consistent than the monolingual English BERT model, especially when these multilingual models are prompted with non-English inputs.

From a model diagnostics perspective, \citet{cao-etal-2021-knowledgeable} found that English PLMs might be biased to making specific predictions based on a predicate's template irrespective of the subjects used to populate this template. Thereafter, \citet{elazar2023measuring} designed a causal framework for modeling multiple co-occurrence statistics that might cause English PLMs to achieve high scores on some of LAMA's predicates.

We focus on why a non-English PLM might fail to recall facts and hypothesize the following possible reasons:

\vspace{-\topsep}
\vspace{-\topsep}
\begin{enumerate}
      \setlength{\parskip}{0pt}
      \setlength{\itemsep}{0pt plus 1pt}    \item The quality of the template might degrade after automatically translating it from English.
    \item Non-English or multilingual PLM are generally pretrained on a lesser amount of non-English data and thus might be less capable of recalling facts efficiently.
    \item Translating the underlying facts of a benchmark, initially designed to probe English PLMs, might cause a representational bias.
\end{enumerate}

\vspace{-\topsep}

While the first two factors are studied in the literature, we believe that the third factor is a major quality issue that previous work has overlooked. Randomly sampling the triples from T-REx might introduce a representation bias toward Western cultures, since only facts aligned to English Wikipedia abstracts are considered. We investigate the presence of such bias (\S\ref{sec:lama_bias}). Moreover, we empirically demonstrate how better model diagnostics can be performed when the benchmark is formed using two diverse and contrasting sets of facts (\S\ref{sec:DLAMA_results}).

 

\section{Cultural Bias in mLAMA}
\label{sec:lama_bias}
Probing PLMs using prompts is an analysis tool attempting to understand how they behave. A biased probing benchmark might be deceiving, as both a good-performing model and a model sharing the same bias found in the benchmark would achieve good performance. In this section, we investigate if the facts within mLAMA might be biased toward Western cultures, which can affect the reliability of the performance scores achieved by PLMs when probed using mLAMA.

\subsection{Quantifying the Cultural Bias}
\label{sec:quantify-bias}
As a proxy for measuring the skewness of the triples of T-REx, LAMA, and X-FACTR toward Western cultures, 26 relation predicates are selected that have a person's name or a place as their subject or object. Moreover, 21 Western countries are identified as representative of Western cultures from Western European and South Western European countries\footnote{According to EuroVoc: \url{https://eur-lex.europa.eu/browse/eurovoc.html?params=72,7206\#arrow_912}}: Andorra, Austria, Belgium, France, Germany, Ireland, Italy, Liechtenstein, Luxembourg, Monaco, Netherlands, Portugal, San Marino, Spain, Switzerland, the United Kingdom, in addition to Canada, the United States of America, Australia, and New Zealand. For each relation predicate out of the 26, triples with a subject or object that either has a country of citizenship or is located in one of the 21 Western countries are counted.

63.6\% of the triples within the LAMA benchmark are related to these Western countries compared to 62.7\% for X-FACTR, and 57.1\% for T-REx (from which LAMA and X-FACTR are sampled)\footnote{Full percentages for each predicate are listed in Table \ref{tab:representation-of-cultures-in-benchmarks}.}. This highlights the issue that aligning Wikidata triples to English Wikipedia abstracts in T-REx would skew them toward Western countries, impacting both LAMA and X-FACTR.

\subsection{Qualitative Analysis of the Bias and its Impact}
\begin{figure*}[!t]
  \centering
  \begin{subfigure}{0.5\linewidth}
      \centering
      \includegraphics[width=\textwidth]{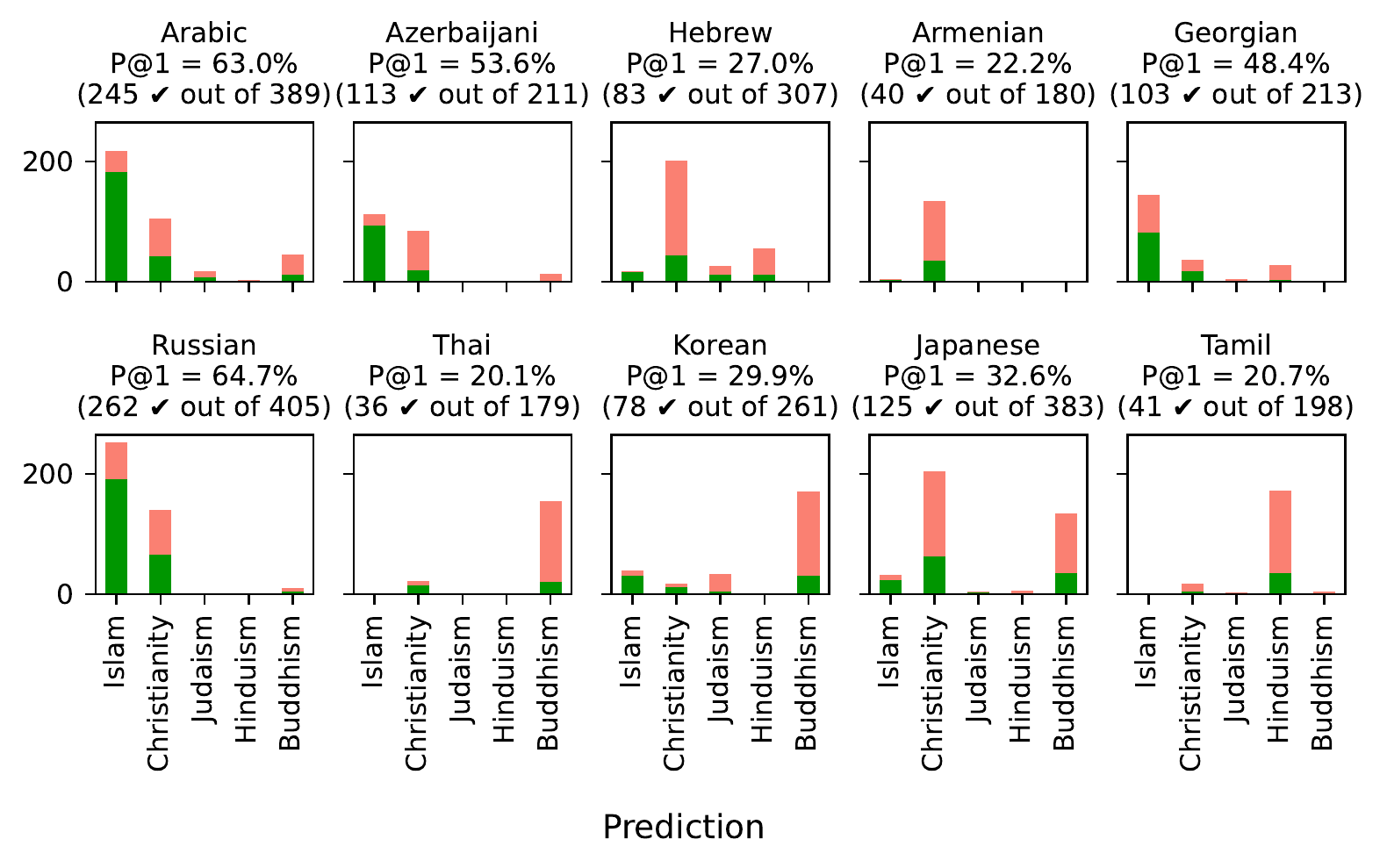}
      \caption{Languages with the least overall performance on mLAMA.}
  \end{subfigure}%
  \begin{subfigure}{0.5\linewidth}
      \centering
      \includegraphics[width=\textwidth]{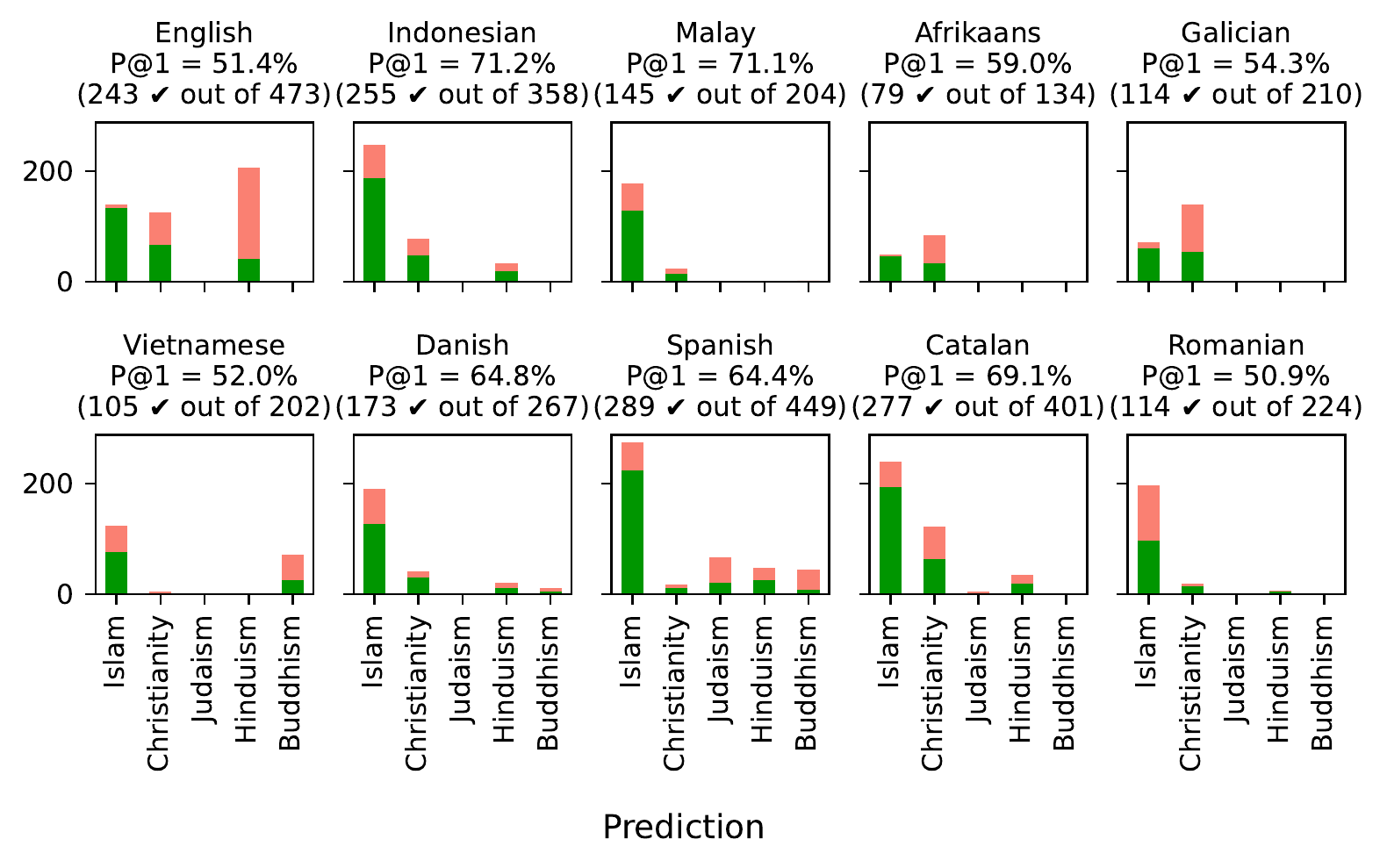}
      \caption{Languages with the highest overall performance on mLAMA.}
  \end{subfigure}
  \caption{The distribution of the predictions of mBERT for the P140 predicate (religion or worldview) for prompts in 20 different languages after merging similar objects' predictions (e.g., Muslim and Islam). The \colorbox{p140_green}{green portion} of the bar represents the triples for which the prediction is correct, while the \colorbox{p140_red}{red portion} represents the triples for which the prediction is wrong. The \textbf{P@1 (Precision at first rank)} metric is the percentage of triples for which the model's first prediction for a triple's subject matches the triple's object.\newline\textbf{Note}: \textbf{P@1} scores are not directly comparable since the number of triples in mLAMA differs between languages.}
  \label{P140-predictions}
\end{figure*}

\citet{kassner-etal-2021-multilingual} used mLAMA to probe mBERT using prompts in 41 languages. We find that all the languages in which prompts achieve the highest performance\footnote{English, Indonesian, Malay, Afrikaans, Galician, Vietnamese, Danish, Spanish, Catalan, Cebuano, Romanian.} use the Latin script, while the ones with the least performance\footnote{Russian, Azerbaijani, Hebrew, Arabic, Korean, Armenian, Georgian, Tamil, Thai, Japanese.} use other scripts. This might be attributed to the model's ability to share cross-lingual representations for common named entities for languages using the Latin script, which allows for cross-lingual knowledge sharing. Moreover, it is known that more than 78\% of mBERT's vocabulary is Latin subwords\footnote{\url{http://juditacs.github.io/2019/02/19/bert-tokenization-stats.html}}.

However, there are still some relation predicates for which a non-Latin scripted language outperforms a Latin-scripted one. The P140\footnote{Wikidata predicates' identifiers format is $P[0-9]+$.} (religion or worldview) predicate is a clear example of these predicates. An example triple for the P140 predicate is: \textbf{(Edward I of England, religion or worldview [P140], Christianity)}. mBERT has higher performance for Arabic (23.1\%), Azerbaijani (8.1\%), Korean (30.1\%), Georgian (35.1\%), Thai (13.4\%), Tamil (4.0\%), Russian (54.6\%), and Japanese (30.0\%) than for English (1.5\%). Looking at the objects for the English triples within mLAMA, we find that 53.7\% of the triples have \textit{Islam} as their object.

While the objects for the P140 predicate should be religions, we find that only seven triples have incorrect inflected forms of \textit{Muslim}, \textit{Christian}, and \textit{Hindu} instead of \textit{Islam}, \textit{Christianity}, and \textit{Hinduism}. Further investigation reveals that the English template used to transform the triples into prompts is (\textit{[X] is affiliated with the [Y] religion .}) which would suit retrieving these infrequent inflected labels than the frequent labels. Therefore, most predictions for the English prompts are considered incorrect justifying the low performance achieved for English. To overcome penalizing these predictions, we mapped the model's predictions and the objects' labels such that for instance \textit{Christian} and \textit{Christianity} are both considered to represent the same prediction \textit{Christianity}, and similarly for \textit{Hinduism} and \textit{Islam}. 

Figure \ref{P140-predictions} shows the distribution of mBERT's predictions for the P140 triples for prompts in 20 different languages after unifying the labels. We observe that: \textbf{(1)} For some languages, the predictions are skewed toward a specific wrong label that is culturally related to these languages. For example, the mode of the predictions of prompts in Armenian, Thai, Korean, and Tamil is Christianity, Buddhism, Buddhism, and Hinduism respectively. \textbf{(2)} Arabic, and Russian prompts tend to yield high performance. The same holds for Indonesian and Malay which achieve similar performance with less skewness in the predictions. Since the label distribution for this predicate within mLAMA is skewed toward a specific label \textit{Islam}, one can not confidently conclude whether the model is choosing the right answer for having some knowledge of the facts or for making a biased guess that luckily coincides with the right label. While these findings signify the possibility that mLAMA is biased for the P140 predicate, it on the other hand might hint that mLAMA is also biased toward Western cultures for most of the remaining predicates. For instance, the P103 (Native Language) predicate in mLAMA has \textit{French} as the correct label for 60.14\% of the triples.

\begin{figure*}[hbtp]
    \centering
    \includegraphics[width=0.8\textwidth]{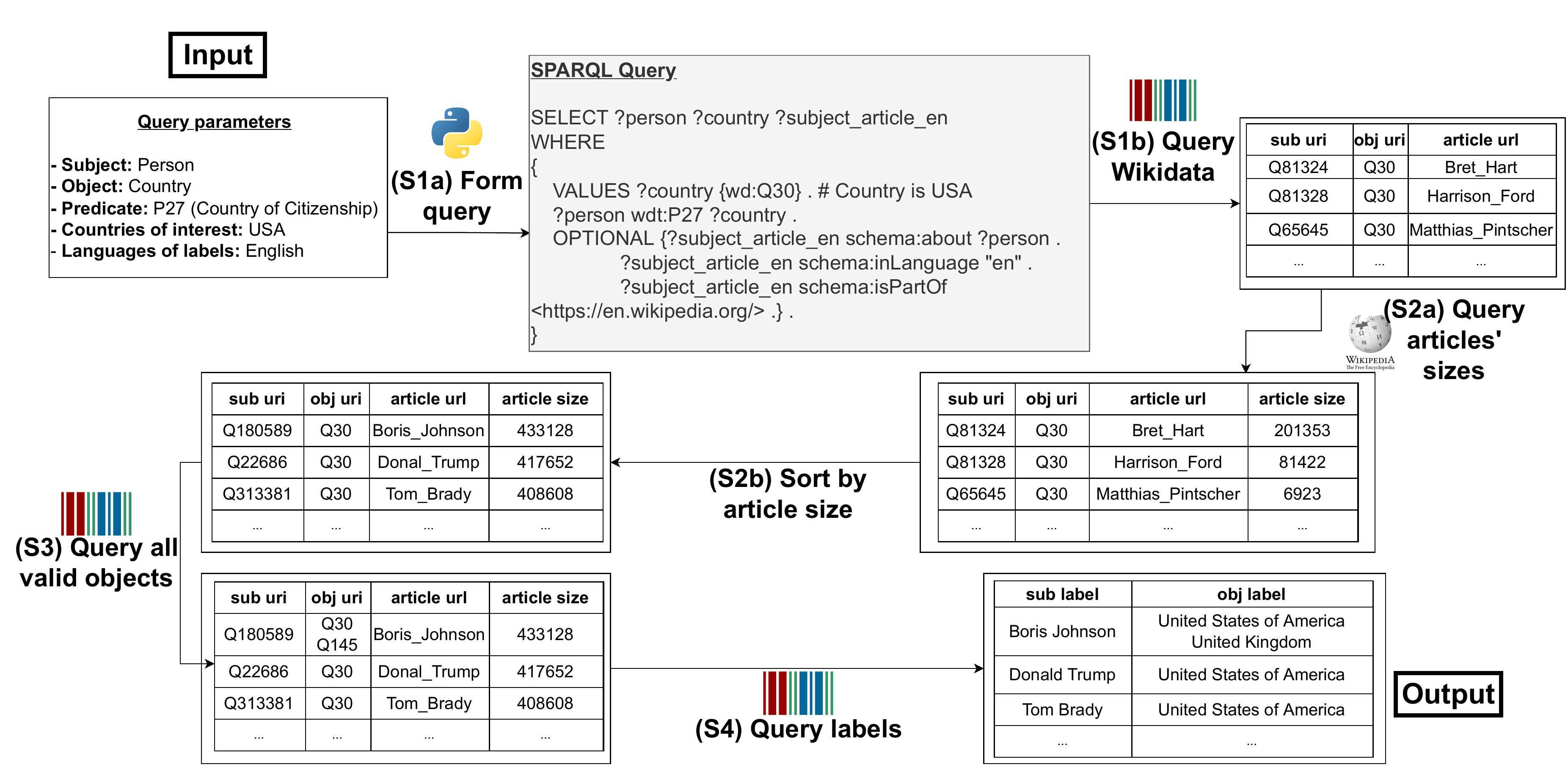}
    \caption{A demonstration of DLAMA's querying framework for the predicate P27 (Country of Citizenship).}
    \label{fig:pipeline}
\end{figure*}

\section{Building DLAMA}
\label{sec:DLAMA}

Our methodology aims at building a culturally diverse benchmark, which would allow for a fairer estimation of a model's capability of memorizing facts. Within DLAMA, query parameters form underlying SPARQL queries that are used to retrieve Wikidata triples as demonstrated in Figure \ref{fig:pipeline}.

To operationalize the concept of cultures, we use countries as a proxy for the cultures of interest. For instance, countries that are members of the Arab League are considered representatives of Arab cultures. Conversely, Western countries mentioned in \S\ref{sec:quantify-bias} represent Western cultures. Furthermore, China, Indonesia, Japan, Malaysia, Mongolia, Myanmar, North Korea, Philippines, Singapore, South Korea, Taiwan, Thailand, and Vietnam are 13 countries from East Asia, and Southeast Asia\footnote{Based on the UN stats classification: \url{https://unstats.un.org/unsd/methodology/m49/}} representing Asian cultures, while Argentina, Bolivia, Brazil, Chile, Colombia, Ecuador, Guyana, Paraguay, Peru, Suriname, Uruguay, Venezuela represent South American cultures.

For predicates in which the subject is a person, we add a filter to the SPARQL query which limits the country of citizenship of the person to a specific set of countries (i.e., a specific culture). For predicates in which the subject is a place, we limit the values of the places to those located in a country within the predefined set of countries related to the target culture.

We implemented a Python interface to simplify the process of querying Wikidata triples. Currently, 20 relation predicates are supported. The user-friendly interface allows the addition of new relation predicates and filters, which we hope would encourage contributions to DLAMA. 

\subsection{Methodology of Querying Triples for a Specific Predicate}
\label{sec:methodology}

\noindent \textbf{Step \#1 - Getting an exhaustive list of triples for a Wikidata predicate}: A set of parameters need to be specified through the Python interface to generate an underlying SPARQL query. These parameters are \textbf{(1)} an entity label for the subject, and an entity label for the object\footnote{\textbf{The used entity labels are} City, Continent, Country, Genre, Instrument, Language, Occupation, Original Network, Person, Piece of Work, Place, Record Label.}, \textbf{(2)} a set of countries representing specific cultures, \textbf{(3)} a Wikidata predicate relating the object to the subject, \textbf{(4)} a list of Wikipedia sites that are expected to contain facts related to each specified country, and \textbf{(5)} a list of languages for which the parallel labels of the subjects and the objects are acquired and later used to populate the multilingual probing templates. In addition to querying the Wikidata Unique Reference Identifiers (URIs) of the subjects and the objects, the Unique Reference Links (URLs) of the Wikipedia articles linked to the subjects are queried as optional fields.

\noindent \textbf{Step \#2 - Sorting the list of retrieved triples by their validity}: Facts on Wikidata are crowdsourced, and contributors are encouraged to add references to the facts they modify. However, lots of the facts on Wikidata still have missing references. Therefore, we use the length of the Wikipedia article corresponding to the triple's subject as a proxy for the validity of the triple. The fact that contributors and editors spent time writing a long Wikipedia article implies that a group of people finds the article important. Therefore they will be keen on making sure the information there is factually sound \cite{bruckman2022should}.
We believe that using the size of the article rather than other metrics such as the number of visits to the Wikipedia article, allows facts related to underrepresented groups on Wikipedia to still be ranked high, thus making the top-ranked facts more diverse and inclusive.
We sort the retrieved triples by the size (in bytes) of the Wikipedia article linked to their subjects. In case a subject has articles on multiple Wikipedia sites, the size of the largest article is used. DLAMA also allows sorting the triples by the total number of edits (revisions) of their subjects' respective articles. 

\noindent \textbf{Step \#3 - Querying all possible objects for each subject}: Since a subject might be linked to multiple objects for the same relation predicate, another query is executed in order to ensure that all these objects are retrieved. For instance, a person might be a citizen of an Arab country in addition to another non-Arab country. This step ensures that the non-Arab country is still considered as a valid country of citizenship for the person, even if the initial query restricted the countries to Arab ones only. While previous benchmarks limited the object for each triple to a single value, we believe it is fairer to allow multiple valid labels instead of randomly picking one label out of the valid ones.


\noindent \textbf{Step \#4 - Querying the labels for the triples}: Till this stage, the subjects and objects are represented by their Wikidata URIs. The Wikidata labels of all the subjects and objects need to be fetched for the languages of interest. Relation triples having missing subject or object labels in any of the languages specified are discarded in order to ensure that the triples are the same for all the languages.

~\\\noindent \textbf{Step \#5 (optional) - Handling overlapping objects}:
The degree of granularity of the objects for Wikidata's relation predicates differs even among triples of the same predicate (e.g.: The official language of \texttt{Australia} is set to \texttt{English} while that of \texttt{The United States of America} is set to \texttt{American English} which is a subclass of \texttt{English}). To avoid penalizing models for picking an object that is a superclass of the correct object, a graph is built, modeling the hierarchical relations between all the objects of the sampled triples of a relation predicate. The graph is later used to augment the valid objects with their superclasses as detailed in \S\ref{sec:aug_step} of the Appendix. 



\subsection{The DLAMA-v1 Benchmark}
We used the above method to build three sets of facts as part of DLAMA-v1 to assess the performance of PLMs on recalling facts related to 21 Western countries as compared to the 22 Arab, 13 Asian countries, and 12 South American countries\footnote{Refer to \S\ref{sec:quantify-bias} and \S\ref{sec:DLAMA} for the list of countries.}. The sets provide examples of how the framework can be used to compile facts from pairs of contrasting cultures. We hope the community will use the framework to introduce new pairs representing other countries and cultures. A maximum of 1000 triples from each predicate out of the 20 supported ones are independently queried for each set of countries within each pair. This ensures that the queried triples are balanced across the two sets of countries within the pair.

In total, the (Arab-West) pair comprises 24535 triples with labels in Arabic and English, as compared to 27076 triples with labels in Korean, and English for the (Asia-West) pair, and 26657 triples with labels in Spanish, and English for the (South America-West) pair. Figure \ref{fig:prompt} shows an example of a triple of DLAMA-v1's (Arab-West) set. The underlying triples belonging to the Western cultures in the 3 sets are not identical. Triples in a set are discarded if their subjects or objects do not have labels in the languages.

Regarding the languages of the labels, Arabic and Korean are chosen as they are two of the least-performing languages on mLAMA. It is expected that facts related to Arab and East Asian/South East Asian countries are relevant to Arabic and Korean PLMs respectively, and would be contrasting to Western facts. Additionally, both languages have non-Latin scripts, use white spaces to separate tokens, and have an inventory of monolingual PLMs. On the other hand, the (South America-West) pair is a trickier case since most South American countries use Spanish as their official language. One can argue that sharing the same language with Spain introduces commonalities between the South-American countries and the Western ones. 

\begin{figure}[tb]
    \centering
    \begin{itemize}
        \itemsep0em 
        \item \scriptsize \textbf{Prompt}: Egypt is located in ...
        \item \textbf{Subject}: \{Egypt\}
        \item \textbf{Set of correct objects}: \{Africa, Asia\}
        \item \textbf{Set of objects of the predicate to be ranked}: \{Africa, Asia, Europe, Insular Oceania, North America\}
    \end{itemize}
    \vspace*{-0.5cm}
    \caption{An example of a prompt created using a relation triple of DLAMA from the P30 (continent) relation predicate for the Arab-Western pair.}
    \label{fig:prompt}
\end{figure}

\noindent\textbf{Overlap between DLAMA-v1 and T-REx}: For the three culture sets, we measured the percentage of triples found in T-REx. 17.92\% of Arab-related facts are in T-REx compared to 39.85\% of Western-related ones in the (Arab-Western) pair. Moreover, 22.64\% of Asian-related facts are found in T-REx compared to 44.43\% of Western-related ones in the (Asia-Western) pair. Lastly, the overlap percentages for the (South America-West) pair are 17.68\% and 32.22\% respectively. These values demonstrate that T-REx has less coverage of the Arab, Asian, and South American factual triples than its coverage of Western triples. Moreover, the fact that T-REx is tuned for higher precision means that its recall is affected and a lot of the Western facts expected to be found in English Wikipedia abstracts are discarded. Conversely, DLAMA-v1 is a less skewed benchmark across different cultures.

\begin{table}[t]
    \begin{subtable}{\linewidth}
    \centering
    \scriptsize
    \resizebox{\columnwidth}{!}{%
       \aboverulesep=0ex 
       \belowrulesep=0ex 
        \begin{tabular}{cc|cc|cc}
        \rule{0pt}{1.1EM}
        \textbf{Prompt} & \textbf{Model} & \multicolumn{2}{c}{$\boldsymbol{P_{@1}}$} & \multicolumn{2}{c}{$\boldsymbol{P_{@1}}$} \\
        \textbf{Lang.} & \textbf{name}& \textbf{Arab} & \textbf{West} & \textbf{DLAMA} &\textbf{mLAMA}\\
        &&$N$=10946&$N$=13589&$N$=24535 & $N$=17128\\
       \midrule
        \rule{0pt}{1.1EM}
       \multirow{2}{*}{Arabic} & mBERT-base &    13.7 &    \textbf{15.1}* &          14.5 & \textbf{15.2}$\dagger$\\   
       & arBERT &    \textbf{33.6}* &    23.0 &          \textbf{27.7}$\dagger$ & 24.4\\
       \midrule
       \multirow{2}{*}{English} & mBERT-base &    21.2 &    \textbf{37.7}* &          30.3 & \textbf{33.9}$\dagger$\\
       & BERT-base &    27.5 &    \textbf{31.3}* &          29.6 & \textbf{37.9}$\dagger$ \\
    \bottomrule
    \end{tabular}%
    }
    \caption{DLAMA-v1 (Arab-West)}
    \end{subtable}

    \begin{subtable}{\linewidth}
    \centering
    \scriptsize
    \resizebox{\columnwidth}{!}{%
           \aboverulesep=0ex 
       \belowrulesep=0ex 
        \begin{tabular}{cc|cc|cc}
        \rule{0pt}{1.1EM}
        \textbf{Prompt} & \textbf{Model} & \multicolumn{2}{c}{$\boldsymbol{P_{@1}}$} & \multicolumn{2}{c}{$\boldsymbol{P_{@1}}$} \\
        \textbf{Lang.} & \textbf{name}& \textbf{Asia} & \textbf{West} & \textbf{DLAMA} &\textbf{mLAMA}\\
    &&$N$=13479&$N$=13588&$N$=27067 & $N$=14217\\
   \midrule
   \multirow{2}{*}{Korean} &  mBERT-base &      16.4 &     \textbf{28.5}* &            \textbf{22.5}$\dagger$ & 15.7\\
   & KyKim &      \textbf{22.1}* &      19.5 &            \textbf{20.8}$\dagger$ & 13.4\\ 
   \midrule
   \multirow{2}{*}{English} & mBERT-base &      33.0 &      \textbf{39.9}* &            \textbf{36.4}$\dagger$ & 35.1\\
   & BERT-base &      \textbf{38.3}* &      31.9 &            35.1 & \textbf{39.0}$\dagger$\\
    \bottomrule
    \end{tabular}
    }
    \caption{DLAMA-v1 (Asia-West)}
    \end{subtable}

    \begin{subtable}{\linewidth}
    \resizebox{\columnwidth}{!}{%
    \aboverulesep=0ex 
    \belowrulesep=0ex 
    \centering
    \scriptsize

    \begin{tabular}{cc|cc|cc}
        \rule{0pt}{1.1EM}
        \textbf{Prompt} & \textbf{Model} & \multicolumn{2}{c}{$\boldsymbol{P_{@1}}$} & \multicolumn{2}{c}{$\boldsymbol{P_{@1}}$} \\
        \textbf{Lang.} & \textbf{name}& \textbf{S. America} & \textbf{West} & \textbf{DLAMA} &\textbf{mLAMA}\\
    && $N$=13071 & $N$=13586& $N$=26657 & $N$=28168\\
   \midrule
   \multirow{2}{*}{Spanish} & mBERT-base &               25.4 &      \textbf{33.8}* &            29.7 & \textbf{30.5}$\dagger$\\
   & BETO &               16.0 &      \textbf{26.5}* &            21.4 & \textbf{22.7}$\dagger$\\ 
   \midrule
   \multirow{2}{*}{English} & mBERT-base &               27.0 &      \textbf{37.6}* &            32.4 & \textbf{33.9}$\dagger$\\
    &  BERT-base &               26.9 &      \textbf{31.3}* &            29.2  & \textbf{37.1}$\dagger$\\
    \bottomrule
    \end{tabular}
    }
    \caption{DLAMA-v1 (South America-West)}
    \end{subtable}

    \caption{Performance of mBERT, and monolingual Arabic (arBERT), Korean (KyKim), Spanish (BETO), and English (BERT-base) language models on the three sets of facts of DLAMA-v1. *: the set of cultures on which a model performs better, $\dagger$: the benchmark on which the model achieves higher $P_{@1}$ score.}
    \label{tab:overall_performance}
\end{table}

\begin{table*}[bthp]
\centering
\scriptsize
\begin{tabular}{lccllll}
\multirow{3}{*}{\textbf{Relation}}&&&\multicolumn{2}{c}{\textbf{Arabic prompts}} & \multicolumn{2}{c}{\textbf{English prompts}}\\
& \multicolumn{2}{c}{$\mathbf{\#\ facts\ (entropy)}$} & \multicolumn{2}{c}{$\mathbf{P_{@1}}$} & \multicolumn{2}{c}{$\mathbf{P_{@1}}$} \\
& $\mathbf{_{Arab}}$ & $\mathbf{_{West}}$ & $\mathbf{_{Arab}}$ & $\mathbf{_{West}}$ & $\mathbf{_{Arab}}$ & $\mathbf{_{West}}$\\
\midrule
                        P17 (Country) &  1000 (3.9) &  1000 (2.8) & \textbf{49.9} &                      47.4 & \textbf{52.2} &                       45.6 \\
                 P19 (Place of birth) &  1000 (3.9) &  1000 (2.6) & \textbf{33.7} &                      22.3 & \textbf{10.1} &                        8.8 \\
                 P20 (Place of death) &  1000 (3.8) &  1000 (2.7) &                      21.3 & \textbf{22.7} &                      14.2 &  \textbf{17.2} \\
         P27 (Country of citizenship) &  1000 (3.8) &  1000 (2.4) & \textbf{38.1} &                      27.9 &                       4.1 &  \textbf{17.5} \\
                      P30 (Continent) &    22 (1.0) &    19 (1.0) & \textbf{45.5} &                      26.3 & \textbf{86.4} &                       84.2 \\
                        P36 (Capital) &    22 (4.5) &    19 (4.2) & \textbf{95.5} &                      78.9 &                      36.4 &  \textbf{84.2} \\
              P37 (Official language) &    22 (0.0) &    19 (2.5) & \textbf{90.9} &                      84.2 &                      95.5 & \textbf{100.0} \\
             P47 (Shares border with) &    22 (2.5) &    19 (2.7) & \textbf{27.3} &                      15.8 &                      68.2 &  \textbf{78.9} \\
               P103 (Native language) &  1000 (1.0) &  1000 (1.7) &                      61.8 & \textbf{72.8} &                      67.7 &  \textbf{74.4} \\
                    P106 (Occupation) &  1000 (2.3) &  1000 (2.0) &  \textbf{3.7} &                       3.3 &                       4.8 &  \textbf{14.3} \\
                         P136 (Genre) &   452 (2.7) &  1000 (2.6) &                       6.6 & \textbf{24.3} &                       4.0 &   \textbf{7.6} \\
                   P190 (Sister city) &    67 (4.9) &   468 (7.3) &                       0.0 &  \textbf{2.6} &  \textbf{6.0} &                        2.8 \\
                  P264 (Record label) &   166 (3.0) &  1000 (5.2) &                       0.0 &  \textbf{0.3} &                       4.2 &   \textbf{7.5} \\
     P364 (Original language of work) &  1000 (0.6) &  1000 (0.4) & \textbf{61.2} &                      48.5 &                      36.1 &  \textbf{88.9} \\
              P449 (Original network) &   127 (4.5) &  1000 (5.3) &  \textbf{0.8} &                       0.4 &                       0.0 &  \textbf{10.8} \\
             P495 (Country of origin) &  1000 (3.1) &  1000 (1.3) & \textbf{18.6} &                       8.7 & \textbf{14.7} &                        5.5 \\
           P530 (Diplomatic relation) &    22 (0.0) &    19 (0.0) &                      22.7 & \textbf{42.1} &                      31.8 &  \textbf{68.4} \\
                   P1303 (Instrument) &  1000 (0.9) &  1000 (1.1) &  \textbf{0.3} &                       0.2 &                       1.9 &  \textbf{27.7} \\
                   P1376 (Capital of) &    24 (4.3) &    26 (4.0) & \textbf{91.7} &                      84.6 & \textbf{79.2} &                       76.9 \\
P1412 (Languages spoken or published) &  1000 (0.8) &  1000 (1.5) & \textbf{67.4} &                      26.1 &                      83.4 &  \textbf{88.7} \\
\midrule
\textbf{Aggregated statistics} & 10946 (2.6) & 13589 (2.7) & \textbf{33.6} &                      23.0 &                      27.5 &  \textbf{31.3} \\

\bottomrule
\end{tabular}
%
\caption{Detailed P@1 scores of arBERT (Arabic prompts) and cased BERT-base (English prompts) on the DLAMA-v1 (Arab-West) set. \textbf{Note:} $\#\ facts$ is the number of facts for each culture within the benchmark, while $(entropy)$ is the entropy of the objects for the facts of each culture.}
\label{tab:detailed_dlama_arab}
\end{table*}

\begin{table*}[thbp]
    \resizebox{\textwidth}{!}{%
    \centering
    \tiny
    \begin{tabular}{lcc}
        \multirow{2}{*}{\textbf{Relation predicate}} & \textbf{Common correct predictions} & \textbf{Common wrong predictions}\\
        & \textbf{(\% of predictions)} & \textbf{(\% of predictions)}\\
        \midrule
        \multicolumn{3}{c}{\textbf{Probing arBERT with \underline{Arabic prompts} populated with Arab facts}}\\
        \midrule

        P17: [X] is located in [Y] . & \colorbox{pred_purple}{\textcolor{white}{Egypt}} (8.4\%) \colorbox{pred_purple}{\textcolor{white}{Algeria}} (7.5\%) \colorbox{pred_purple}{\textcolor{white}{Morocco}} (5.6\%) & \colorbox{pred_purple}{\textcolor{white}{Morocco}} (10.5\%) \colorbox{pred_blue_bell}{\textcolor{white}{Turkey}} (7.1\%) \colorbox{pred_purple}{\textcolor{white}{Tunisia}} (6.7\%) \\
        P19: [X] was born in [Y] . & \colorbox{pred_purple}{\textcolor{white}{Egypt}} (9.0\%) \colorbox{pred_purple}{\textcolor{white}{Algeria}} (4.2\%) \colorbox{pred_purple}{\textcolor{white}{Morocco}} (3.9\%) & \colorbox{pred_purple}{\textcolor{white}{Algeria}} (18.8\%) \colorbox{pred_purple}{\textcolor{white}{Tunisia}} (7.2\%) \colorbox{pred_purple}{\textcolor{white}{Morocco}} (5.9\%) \\
        P20: [X] died in [Y] . & \colorbox{pred_purple}{\textcolor{white}{Egypt}} (9.8\%) \colorbox{pred_purple}{\textcolor{white}{Baghdad}} (2.0\%) \colorbox{pred_purple}{\textcolor{white}{Tunisia}} (1.1\%) & \colorbox{pred_dark_orange}{Paris} (28.7\%) \colorbox{pred_purple}{\textcolor{white}{Egypt}} (8.8\%) \colorbox{pred_purple}{\textcolor{white}{Tunisia}} (5.7\%) \\
        P27: [X] is [Y] citizen . & \colorbox{pred_purple}{\textcolor{white}{Saudi Arabia}} (4.5\%) \colorbox{pred_purple}{\textcolor{white}{Morocco}} (4.2\%) \colorbox{pred_purple}{\textcolor{white}{Egypt}} (3.8\%) & \colorbox{pred_purple}{\textcolor{white}{State of Palestine}} (15.2\%) \colorbox{pred_purple}{\textcolor{white}{Republic of Egypt}} (9.3\%) \colorbox{pred_purple}{\textcolor{white}{Iraqi Republic}} (5.2\%) \\
        P495: [X] was created in [Y] . & \colorbox{pred_purple}{\textcolor{white}{Egypt}} (6.4\%) \colorbox{pred_purple}{\textcolor{white}{Morocco}} (3.0\%) \colorbox{pred_purple}{\textcolor{white}{France}} (2.9\%) & \colorbox{pred_dark_orange}{France} (26.9\%) \colorbox{pred_dark_orange}{Germany} (19.5\%) \colorbox{pred_purple}{\textcolor{white}{Morocco}} (10.2\%) \\
        P103: The native language of [X] is [Y] . & \colorbox{pred_purple}{\textcolor{white}{Arabic}} (59.7\%) \colorbox{pred_dark_orange}{French} (1.1\%) \colorbox{pred_dark_orange}{English} (0.4\%) & \colorbox{pred_dark_orange}{English} (8.8\%) \colorbox{pred_purple}{\textcolor{white}{Arabic}} (8.6\%) \colorbox{pred_purple}{\textcolor{white}{Shilha}} (8.4\%) \\
        P364: The original language of [X] is [Y] . & \colorbox{pred_purple}{\textcolor{white}{Arabic}} (58.9\%) \colorbox{pred_dark_orange}{French} (1.1\%) \colorbox{pred_dark_orange}{English} (0.9\%) & \colorbox{pred_purple}{\textcolor{white}{Shilha}} (11.7\%) \colorbox{pred_dark_orange}{English} (8.5\%) \colorbox{pred_dark_orange}{French} (7.6\%) \\
        P1412: [X] used to communicate in [Y] . & \colorbox{pred_purple}{\textcolor{white}{Arabic}} (62.5\%) \colorbox{pred_dark_orange}{French} (4.7\%) \colorbox{pred_dark_orange}{Spanish} (0.1\%) & \colorbox{pred_purple}{\textcolor{white}{Syrian Arabic}} (15.1\%) \colorbox{pred_purple}{\textcolor{white}{Arabic}} (9.1\%) \colorbox{pred_dark_orange}{French} (5.0\%) \\

        \midrule
        \multicolumn{3}{c}{\textbf{Probing arBERT with \underline{Arabic prompts} populated with Western facts}}\\
        \midrule

        P17: [X] is located in [Y] . & \colorbox{pred_purple}{\textcolor{white}{France}} (13.5\%) \colorbox{pred_purple}{\textcolor{white}{United States of America}} (9.3\%) \colorbox{pred_purple}{\textcolor{white}{Spain}} (9.1\%) & \colorbox{pred_purple}{\textcolor{white}{Germany}} (8.3\%) \colorbox{pred_light_orange}{South Africa} (7.5\%) \colorbox{pred_purple}{\textcolor{white}{France}} (6.6\%) \\
        P19: [X] was born in [Y] . & \colorbox{pred_purple}{\textcolor{white}{Germany}} (7.4\%) \colorbox{pred_purple}{\textcolor{white}{Italy}} (4.6\%) \colorbox{pred_purple}{\textcolor{white}{New York City}} (3.5\%) & \colorbox{pred_purple}{\textcolor{white}{New York City}} (26.0\%) \colorbox{pred_purple}{\textcolor{white}{Germany}} (18.0\%) \colorbox{pred_purple}{\textcolor{white}{Italy}} (10.8\%) \\
        P20: [X] died in [Y] . & \colorbox{pred_purple}{\textcolor{white}{Paris}} (9.0\%) \colorbox{pred_purple}{\textcolor{white}{Germany}} (3.5\%) \colorbox{pred_purple}{\textcolor{white}{Italy}} (3.4\%) & \colorbox{pred_purple}{\textcolor{white}{Paris}} (33.4\%) \colorbox{pred_purple}{\textcolor{white}{New York City}} (19.7\%) \colorbox{pred_purple}{\textcolor{white}{London}} (7.6\%) \\
        P27: [X] is [Y] citizen . & \colorbox{pred_purple}{\textcolor{white}{France}} (8.7\%) \colorbox{pred_purple}{\textcolor{white}{Germany}} (6.5\%) \colorbox{pred_purple}{\textcolor{white}{United States of America}} (6.3\%) & \colorbox{pred_purple}{\textcolor{white}{Germany}} (12.9\%) \colorbox{pred_dark_orange}{French protectorate of Tunisia} (11.0\%) \colorbox{pred_purple}{\textcolor{white}{Republic of Ireland}} (6.8\%) \\
        P495: [X] was created in [Y] . & \colorbox{pred_purple}{\textcolor{white}{United States of America}} (3.9\%) \colorbox{pred_purple}{\textcolor{white}{France}} (2.4\%) \colorbox{pred_purple}{\textcolor{white}{Germany}} (1.7\%) & \colorbox{pred_purple}{\textcolor{white}{Germany}} (44.2\%) \colorbox{pred_purple}{\textcolor{white}{France}} (23.4\%) \colorbox{pred_dark_orange}{Algeria} (4.5\%) \\
        P103: The native language of [X] is [Y] . & \colorbox{pred_purple}{\textcolor{white}{English}} (53.1\%) \colorbox{pred_purple}{\textcolor{white}{French}} (12.8\%) \colorbox{pred_purple}{\textcolor{white}{German}} (3.2\%) & \colorbox{pred_purple}{\textcolor{white}{French}} (7.7\%) \colorbox{pred_purple}{\textcolor{white}{English}} (4.2\%) \colorbox{pred_purple}{\textcolor{white}{Spanish}} (3.9\%) \\
        P364: The original language of [X] is [Y] . & \colorbox{pred_purple}{\textcolor{white}{English}} (47.2\%) \colorbox{pred_purple}{\textcolor{white}{French}} (0.7\%) \colorbox{pred_purple}{\textcolor{white}{German}} (0.2\%) & \colorbox{pred_dark_orange}{Arabic} (30.8\%) \colorbox{pred_purple}{\textcolor{white}{French}} (5.7\%) \colorbox{pred_dark_orange}{Shilha} (5.1\%) \\
        P1412: [X] used to communicate in [Y] . & \colorbox{pred_purple}{\textcolor{white}{French}} (12.6\%) \colorbox{pred_purple}{\textcolor{white}{German}} (7.7\%) \colorbox{pred_purple}{\textcolor{white}{Spanish}} (3.4\%) & \colorbox{pred_dark_orange}{Arabic} (55.9\%) \colorbox{pred_purple}{\textcolor{white}{German}} (8.3\%) \colorbox{pred_purple}{\textcolor{white}{French}} (3.4\%) \\
        
        \midrule
        
        \multirow{2}{*}{\textbf{Relation predicate}} & \textbf{Common correct predictions} & \textbf{Common wrong predictions}\\
        & \textbf{(\% of predictions)} & \textbf{(\% of predictions)}\\
        \midrule
        \multicolumn{3}{c}{\textbf{Probing BERT-base with \underline{English prompts} populated with Arab facts}}\\
        \midrule
        P17: [X] is located in [Y] . & \colorbox{pred_purple}{\textcolor{white}{Algeria}} (9.0\%) \colorbox{pred_purple}{\textcolor{white}{Egypt}} (8.6\%) \colorbox{pred_purple}{\textcolor{white}{Iraq}} (4.6\%) & \colorbox{pred_purple}{\textcolor{white}{Bahrain}} (8.6\%) \colorbox{pred_light_orange}{Moscow} (4.1\%) \colorbox{pred_purple}{\textcolor{white}{Lebanon}} (3.8\%) \\
        P19: [X] was born in [Y] . & \colorbox{pred_purple}{\textcolor{white}{Cairo}} (3.8\%) \colorbox{pred_purple}{\textcolor{white}{Baghdad}} (3.0\%) \colorbox{pred_purple}{\textcolor{white}{Damascus}} (0.5\%) & \colorbox{pred_purple}{\textcolor{white}{Baghdad}} (31.1\%) \colorbox{pred_purple}{\textcolor{white}{Cairo}} (18.1\%) \colorbox{pred_dark_orange}{Paris} (6.4\%) \\
        P20: [X] died in [Y] . & \colorbox{pred_purple}{\textcolor{white}{Cairo}} (10.9\%) \colorbox{pred_purple}{\textcolor{white}{Baghdad}} (2.0\%) \colorbox{pred_purple}{\textcolor{white}{Egypt}} (0.7\%) & \colorbox{pred_purple}{\textcolor{white}{Cairo}} (45.9\%) \colorbox{pred_dark_orange}{Paris} (19.2\%) \colorbox{pred_purple}{\textcolor{white}{Baghdad}} (8.0\%) \\
        P27: [X] is [Y] citizen . & \colorbox{pred_dark_orange}{France} (1.8\%) \colorbox{pred_purple}{\textcolor{white}{Qatar}} (1.3\%) \colorbox{pred_blue_bell}{\textcolor{white}{Israel}} (0.4\%) & \colorbox{pred_purple}{\textcolor{white}{Qatar}} (73.9\%) \colorbox{pred_blue_bell}{\textcolor{white}{Pakistan}} (8.8\%) \colorbox{pred_blue_bell}{\textcolor{white}{Israel}} (2.8\%) \\
        P495: [X] was created in [Y] . & \colorbox{pred_purple}{\textcolor{white}{Egypt}} (10.0\%) \colorbox{pred_purple}{\textcolor{white}{Algeria}} (1.1\%) \colorbox{pred_purple}{\textcolor{white}{Iraq}} (0.7\%) & \colorbox{pred_light_orange}{Japan} (25.2\%) \colorbox{pred_blue_bell}{\textcolor{white}{India}} (12.2\%) \colorbox{pred_purple}{\textcolor{white}{Egypt}} (9.2\%) \\
        P103: The native language of [X] is [Y] . & \colorbox{pred_purple}{\textcolor{white}{Arabic}} (66.2\%) \colorbox{pred_dark_orange}{French} (0.8\%) \colorbox{pred_dark_orange}{English} (0.4\%) & \colorbox{pred_purple}{\textcolor{white}{Arabic}} (12.2\%) \colorbox{pred_blue_bell}{\textcolor{white}{Urdu}} (6.5\%) \colorbox{pred_purple}{\textcolor{white}{Kurdish}} (4.3\%) \\
        P364: The original language of [X] is [Y] . & \colorbox{pred_purple}{\textcolor{white}{Arabic}} (30.7\%) \colorbox{pred_dark_orange}{English} (3.5\%) \colorbox{pred_dark_orange}{French} (1.6\%) & \colorbox{pred_dark_orange}{English} (44.6\%) \colorbox{pred_dark_orange}{French} (4.3\%) \colorbox{pred_blue_bell}{\textcolor{white}{Hindi}} (2.1\%) \\
        P1412: [X] used to communicate in [Y] . & \colorbox{pred_purple}{\textcolor{white}{Arabic}} (78.3\%) \colorbox{pred_dark_orange}{English} (2.8\%) \colorbox{pred_dark_orange}{French} (1.8\%) & \colorbox{pred_purple}{\textcolor{white}{Arabic}} (8.5\%) \colorbox{pred_dark_orange}{English} (4.5\%) \colorbox{pred_blue_bell}{\textcolor{white}{Urdu}} (1.2\%) \\        

        \midrule    
        \multicolumn{3}{c}{\textbf{Probing BERT-base with \underline{English prompts} populated with Western facts}}\\
        \midrule

        P17: [X] is located in [Y] . & \colorbox{pred_purple}{\textcolor{white}{France}} (15.7\%) \colorbox{pred_purple}{\textcolor{white}{Spain}} (10.3\%) \colorbox{pred_purple}{\textcolor{white}{Germany}} (8.2\%) & \colorbox{pred_blue_bell}{\textcolor{white}{Georgia}} (10.1\%) \colorbox{pred_blue_bell}{\textcolor{white}{Moscow}} (9.1\%) \colorbox{pred_purple}{\textcolor{white}{Canada}} (6.0\%) \\
        P19: [X] was born in [Y] . & \colorbox{pred_purple}{\textcolor{white}{Paris}} (3.4\%) \colorbox{pred_purple}{\textcolor{white}{Berlin}} (0.9\%) \colorbox{pred_purple}{\textcolor{white}{London}} (0.7\%) & \colorbox{pred_purple}{\textcolor{white}{Chicago}} (25.4\%) \colorbox{pred_purple}{\textcolor{white}{London}} (22.1\%) \colorbox{pred_purple}{\textcolor{white}{Paris}} (9.5\%) \\
        P20: [X] died in [Y] . & \colorbox{pred_purple}{\textcolor{white}{Paris}} (9.4\%) \colorbox{pred_purple}{\textcolor{white}{London}} (2.7\%) \colorbox{pred_purple}{\textcolor{white}{Rome}} (2.6\%) & \colorbox{pred_purple}{\textcolor{white}{Paris}} (29.0\%) \colorbox{pred_purple}{\textcolor{white}{London}} (22.9\%) \colorbox{pred_purple}{\textcolor{white}{Rome}} (8.3\%) \\
        P27: [X] is [Y] citizen . & \colorbox{pred_purple}{\textcolor{white}{France}} (11.0\%) \colorbox{pred_purple}{\textcolor{white}{Italy}} (2.5\%) \colorbox{pred_purple}{\textcolor{white}{Canada}} (1.1\%) & \colorbox{pred_purple}{\textcolor{white}{British America}} (44.5\%) \colorbox{pred_purple}{\textcolor{white}{Austria}} (8.5\%) \colorbox{pred_purple}{\textcolor{white}{Canada}} (6.8\%) \\
        P495: [X] was created in [Y] . & \colorbox{pred_purple}{\textcolor{white}{France}} (2.2\%) \colorbox{pred_purple}{\textcolor{white}{Germany}} (1.4\%) \colorbox{pred_light_orange}{Japan} (0.6\%) & \colorbox{pred_light_orange}{Japan} (61.3\%) \colorbox{pred_purple}{\textcolor{white}{England}} (10.1\%) \colorbox{pred_light_orange}{India} (4.6\%) \\
        P103: The native language of [X] is [Y] . & \colorbox{pred_purple}{\textcolor{white}{English}} (50.8\%) \colorbox{pred_purple}{\textcolor{white}{French}} (13.3\%) \colorbox{pred_purple}{\textcolor{white}{German}} (3.3\%) & \colorbox{pred_purple}{\textcolor{white}{Spanish}} (6.8\%) \colorbox{pred_purple}{\textcolor{white}{German}} (3.6\%) \colorbox{pred_purple}{\textcolor{white}{French}} (3.2\%) \\
        P364: The original language of [X] is [Y] . & \colorbox{pred_purple}{\textcolor{white}{English}} (85.4\%) \colorbox{pred_purple}{\textcolor{white}{French}} (1.5\%) \colorbox{pred_purple}{\textcolor{white}{German}} (0.9\%) & \colorbox{pred_purple}{\textcolor{white}{Latin}} (2.1\%) \colorbox{pred_purple}{\textcolor{white}{English}} (2.0\%) \colorbox{pred_purple}{\textcolor{white}{French}} (1.9\%) \\
        P1412: [X] used to communicate in [Y] . & \colorbox{pred_purple}{\textcolor{white}{English}} (59.8\%) \colorbox{pred_purple}{\textcolor{white}{French}} (13.1\%) \colorbox{pred_purple}{\textcolor{white}{German}} (7.3\%) & \colorbox{pred_purple}{\textcolor{white}{English}} (3.6\%) \colorbox{pred_purple}{\textcolor{white}{Spanish}} (2.2\%) \colorbox{pred_dark_orange}{Arabic} (1.2\%) \\
    \bottomrule
    \end{tabular}
    }
    \caption{The most common predictions for monolingual arBERT and BERT-base models when probed by DLAMA-v1 (Arab-West) with English and Arabic prompts respectively. \colorbox{pred_purple}{\textcolor{white}{Purple}} culturally related prediction, \colorbox{pred_blue_bell}{\textcolor{white}{Blue bell}} culturally proximate prediction, \colorbox{pred_light_orange}{Light Orange} culturally proximate prediction to another culture, \colorbox{pred_dark_orange}{Orange} culturally related prediction to the other culture. \textbf{Note:} The Arabic prompts/entities are translated for clarity.}
    \label{tab:language_bias_arBERT}
\end{table*}

\section{Probing PLMs via DLAMA-v1}
\label{sec:DLAMA_results}
\subsection{Experimental Setup}
We follow mLAMA's probing setup to evaluate the PLMs' factual knowledge. For each relation predicate [PREDICATE], the set $\{$OBJECTS$\}$ of unique objects of the triples is first compiled. Then, for each relation triple within the [PREDICATE], the PLM is asked to assign a score for each object within $\{$OBJECTS$\}$ by computing the probability of having this object replacing the masked tokens. This setup asks the model to choose the correct answer out of a set of possible choices, instead of decoding the answer as a generation task. The templates used in DLAMA to convert triples into natural language prompts are adapted from mLAMA and listed in Table \ref{tab:mLAMA_templates} of the Appendix.

\noindent\textbf{Models}: We evaluated the cased multilingual BERT-base, and the cased English BERT-base using all the sets of facts of DLAMA-v1. Moreover, a monolingual Arabic BERT-base model \textbf{arBERT} \cite{abdul-mageed-etal-2021-arbert}, a monolingual Korean BERT-base model \textbf{KyKim BERT-base} \cite{kim2020lmkor}, and a monolingual cased Spanish BERT-base model \textbf{BETO} \cite{CaneteCFP2020} are evaluated using the (Arab-West), the (Asia-West), and the (South America-West) pairs respectively. We focus on BERT models to compare our results to those previously reported on mLAMA.

\subsection{Aggregated Results}
Precision at the first rank (P@1) is the metric used to evaluate the performance of the models. P@1 is the percentage of triples for which the first prediction of the model matches one of the objects for this triple. In order to quantify the diversity of the objects of a relation predicate for each culture, an entropy score is computed. For each triple of a relation predicate, only the most frequent object among the list of valid objects is considered. The entropy score is computed as $Entropy(\{objs\}) = \sum_{o \in \{objs\}}{-p_{o} * \log(p_{o})}$; where $p_{o}$ is the probability of object $o$ across the set of objects $\{objs\}$. The higher the entropy of the objects is, the more diverse the objects are, and thus the harder the predicate would be for a model to randomly achieve high (P@1) scores.

Looking at the performance of models on DLAMA indicated in Table \ref{tab:overall_performance}, \textbf{(1)} we find how the facts' relevance to the probed model's language affects the results. For instance, arBERT and KyKim perform better on non-Western facts than on Western ones. Conversely, the English BERT-base model performs better on Western facts for the (Arab-West) pair. The same observation tends to hold for individual predicates as shown in Table \ref{tab:detailed_dlama_arab}. \textbf{(2)} Moreover, arBERT and KyKim achieve lower performance on mLAMA than their performance on DLAMA-v1, while the English BERT-base and BETO models achieve higher $P_{@1}$ scores on mLAMA than on DLAMA-v1. This is expected given the bias mLAMA has toward facts from Western cultures.

\subsection{Revisiting the Language bias of PLMs}
\label{sec:language_bias}

\citet{kassner-etal-2021-multilingual} showed that for prompts in English, German, Dutch, and Italian, mBERT is biased toward predicting the language or the country name related to the language of the prompts (e.g., Filling the masked object with \textit{Italy} if the prompt's language is \textit{Italian}). This phenomenon is not a bias if most of the triples in the underlying subset of mLAMA for a language are also biased toward the same label. For DLAMA, looking at the P@1 scores in Table \ref{tab:detailed_dlama_arab} in addition to checking the most common predictions of arBERT and the cased BERT-base models in Table \ref{tab:language_bias_arBERT} provides a better diagnostic tool for analyzing the models' behavior\footnote{A similar analysis for the other two sets of contrasting cultures can be found in \S\ref{sec:qualitative-analysis-other-models} of the Appendix.}. For the P364 predicate, the models perform better on their culturally proximate triples. This can be attributed to the Language bias phenomenon which is indicated by arBERT predicting \textit{Arabic} for 30.8\% of Western facts, while BERT-base predicting \textit{English} for 44.6\% of Arab facts. On the other hand, both models achieve high P@1 scores for P17 and P103. Even when the models make wrong predictions for triples of these predicates, the predictions can be considered to be educated guesses, as they are still relevant to the culture to which the triples belong. Lastly, the models perform poorly on P495 for being biased toward specific objects irrespective of the culture of the triples (\textit{Japan} for BERT-base, \textit{Germany} and \textit{France} for arBERT). These three patterns can be noticed thanks to having a contrastive set of facts representing two different cultures.


\subsection{Pilot Evaluation for a Large Language Model}
\label{sec:gpt3.5}
Given the success of large language models (LLMs) (\citealp{brown2020language}; \citealp{scao2022bloom}), we evaluated the performance of the GPT3.5-turbo model on tuples from the P30, P36, P37, P47, P103, P530, and P1376 predicates of DLAMA-v1 (Arab-West). To probe the model, the Arabic and English templates for these predicates were mapped into questions listed in Table \ref{tab:GPT3.5_questions}. While the model is instructed to only respond with an entity, it sometimes provides a full sentence. Consequently, we consider the model's response to a question to be correct if one of the valid objects of the tuple used to populate the question is a substring of the response. GPT3.5's probing setup is harder than BERT's setup in which an answer is chosen from a set of unique objects for the predicate. Nevertheless, GPT3.5 achieves superior performance compared to the monolingual BERT models as per Table \ref{tab:GPT3.5}. However, GPT3.5 seems to be hallucinating for a lot of the tuples within the P190 (Sister City) predicate (e.g.: \textbf{The twin city of Nice is Naples.}). Such issues might be unnoticed unless benchmarks like DLAMA are used to systematically evaluate the LLMs.

\section{Conclusion}
\label{conclusion}
Previous work suggested that English prompts are more capable of recalling facts from multilingual pretrained language models. We show that the facts within the underlying probing benchmark (mLAMA) are skewed toward Western countries, which makes them more relevant to English. Hence, we propose a new framework (DLAMA) that permits the curation of culturally diverse facts directly from Wikidata. Three new sets of facts are released as part of the DLAMA-v1 benchmark containing factual triples representing 20 relation predicates comprising facts from (Arab-Western), (Asian-Western), and (South American-Western) countries, with a more balanced representation between the countries within each pair. The results of probing PLMs on the DLAMA-v1 support that mBERT has a better performance recalling Western facts than non-Western ones irrespective of the prompt's language. Monolingual Arabic and Korean models on the other hand perform better on culturally proximate facts. We believe the probing results are more trustable and fairer when the underlying benchmark is less skewed toward specific countries, languages, or cultures. Moreover, we find that even when the model's prediction does not match any of the correct labels, the model might be making an educated guess relevant to the culture of the underlying facts. This finding augments previous experiments which showed that models tend to have a language bias, by which a model tends to overgenerate a specific prediction for each prompting language irrespective of the triple's subject used to fill in the prompt. Finally, our framework is open-sourced for the community to contribute new pairs to the DLAMA benchmark in the future.

\section*{Limitations}
We acknowledge that the methodology used to build DLAMA-v1 still has limitations related to the information within its relation triples. While directly querying Wikidata as a dynamic source of facts provides the flexibility needed to acquire data that is relevant to different cultures (as opposed to using the static T-REx dump of triples), the diversity of the triples that are compiled depends on the availability of a diverse set of facts on Wikidata in the first place. For instance, the smaller number of relation triples related to Arab countries for the predicates (P136 - Genre), (P190 - Sister city), and (P449 - Original network) in DLAMA-v1 (Arab-West) demonstrates the difficulty of querying the exact number of facts for both cultures despite using exactly the same queries with the only difference being limiting the region to which the triples belong.
Another limitation is the inability to enumerate valid and fine-grained subclasses of objects for specific subjects, if these fine-grained objects are not on Wikidata. Steps \#3 and \#5 of DLAMA explained in \S\ref{sec:methodology} ensure that a possible and more general object is still valid for a specific subject. However, inferring a more specified object from a generic one is impossible. For example, the fact that someone speaks ``American English" implies that they speak English as well, but knowing that someone speaks ``English" is not enough to speculate about their dialect (i.e.: ``American English", ``British English", etc.).

While the triples within DLAMA are sampled by picking the ones whose subjects have the largest Wikipedia articles' sizes, the infeasibility of manually reviewing the large number of diverse facts within DLAMA-v1 makes it hard to claim that the facts are free of inaccuracies or missing information. More broadly, DLAMA supports relations predicates that are already part of mLAMA to fairly compare the results on DLAMA to those previously reported on mLAMA. 
Moreover, we make sure that the subjects and the objects of the relation triples are available in the different languages of interest. Having these constraints might imply that some culturally relevant facts might have been dropped out of DLAMA-v1 (e.g., Predicates that are not part of mLAMA, or triples having missing labels in one of the languages of interest).

Lastly, we used mLAMA's probing setup in which the models rank a predefined set of objects for each prompt. Their prediction is correct if the top-ranked object is one of the valid labels for the corresponding relation triple used to populate the prompt. Therefore, a model's performance is expected to be higher than that achieved by a generative setup in which the model is asked to generate the most probable completions for the masked tokens.

\section*{Ethics Statement}
We believe that using a set of countries to represent cultures is just a proxy for acquiring a more diverse set of facts that are less skewed toward a specific culture. More specifically, using the terms Arab cultures, Western cultures, and Asian cultures simplifies the differences between the cultures within the countries that we have used to represent these macro-cultures. On the other hand, we still think that the differences between Asian cultures are less subtle than between them and Western cultures.

We also acknowledge that the accuracy and validity of some relation triples queried from Wikidata might be biased by the views of the people who added such information to Wikidata. This might be particularly vibrant for relation triples related to zones with political/ sectarian wars and conflicts.

\section*{Acknowledgments}
This work was supported in part by the UKRI Centre for Doctoral Training in Natural Language
Processing, funded by the UKRI (grant EP/S022481/1) and the University of Edinburgh, School
of Informatics. Amr is grateful to Matthias Lindemann for recommending Wikidata, Aida Tarighat for the early discussions about the benchmark, Laurie Burchell, Bálint Gyevnár, and Shangmin Guo for reviewing the manual prompts, Coleman Haley for the multiple discussions about the figures, Anna Kapron-King and Gautier Dagan for proof-reading the abstract, and lastly, Dilara Keküllüo\u{g}lu and Björn Ross for their valuable reviews of the paper's final draft.


\bibliography{anthology,custom}
\bibliographystyle{acl_natbib}

\appendix
\setcounter{table}{0}
\setcounter{figure}{0}
\renewcommand{\thetable}{\Alph{section}\arabic{table}}
\renewcommand{\thefigure}{\Alph{section}\arabic{figure}}

\section{Detailed Bias Values within the Factual Knowledge Benchmarks}
\label{sec:appendix}
\begin{sidewaystable*}[htbp]
   \aboverulesep=0ex 
   \belowrulesep=0ex 
    \centering
    \resizebox{\textwidth}{!}{%
    \begin{tabular}{lcc|cc|cc}
    \rule{0pt}{1.1EM}

    \multirow{2}{*}{\textbf{Wikidata predicate}}&\multicolumn{2}{c}{\textbf{T-REx}}&\multicolumn{2}{c}{\textbf{LAMA}}&\multicolumn{2}{c}{\textbf{X-FACTR}}\\
    &Western countries &Rest of the world &Western countries &Rest of the world&Western countries &Rest of the world \\
    \midrule
    \rule{0pt}{1.1EM}
    P17 (Country) &321988 (44.0\%) & \textbf{\underline{410512 (56.0\%)}} &386 (41.6\%) & \textbf{\underline{542 (58.4\%)}} &457 (45.7\%) & \textbf{\underline{543 (54.3\%)}} \\
    P19 (Place of birth) &\textbf{\underline{156579 (68.2\%)}} &73069 (31.8\%) &\textbf{\underline{730 (77.4\%)}} &213 (22.6\%) &\textbf{\underline{680 (68.0\%)}} &320 (32.0\%) \\
    P20 (Place of death) &\textbf{\underline{63250 (76.0\%)}} &19962 (24.0\%) &\textbf{\underline{734 (77.2\%)}} &217 (22.8\%) &\textbf{\underline{757 (75.7\%)}} &243 (24.3\%) \\
    P27 (Country of citizenship) &\textbf{\underline{253402 (63.8\%)}} &143837 (36.2\%) &404 (41.9\%) & \textbf{\underline{560 (58.1\%)}} &\textbf{\underline{623 (62.3\%)}} &377 (37.7\%) \\
    P36 (Capital) &4011 (44.8\%) &\textbf{\underline{4936 (55.2\%)}} &\textbf{\underline{436 (62.0\%)}} &267 (38.0\%) &418 (41.8\%) & \textbf{\underline{582 (58.2\%)}} \\
    P39 (Position held) &\textbf{\underline{7610 (50.1\%)}} &7581 (49.9\%) &380 (42.6\%) & \textbf{\underline{512 (57.4\%)}} &\textbf{\underline{504 (50.4\%)}} &496 (49.6\%) \\
    P47 (Shares border with) &\textbf{\underline{29427 (76.6\%)}} &9010 (23.4\%) &\textbf{\underline{529 (57.4\%)}} &393 (42.6\%) &\textbf{\underline{762 (76.2\%)}} &238 (23.8\%) \\
    P101 (Field of work) &\textbf{\underline{3396 (54.1\%)}} &2885 (45.9\%) &\textbf{\underline{365 (52.5\%)}} &330 (47.5\%) &\textbf{\underline{539 (53.9\%)}} &461 (46.1\%) \\
    P103 (Native language) &\textbf{\underline{6983 (78.4\%)}} &1926 (21.6\%) &\textbf{\underline{778 (79.7\%)}} &198 (20.3\%) &\textbf{\underline{787 (78.7\%)}} &213 (21.3\%) \\
    P106 (Occupation) &\textbf{\underline{203644 (58.7\%)}} &143177 (41.3\%) &\textbf{\underline{631 (65.9\%)}} &327 (34.1\%) &\textbf{\underline{578 (57.8\%)}} &422 (42.2\%) \\
    P108 (Employer) &\textbf{\underline{27119 (91.2\%)}} &2605 (8.8\%) &\textbf{\underline{371 (96.9\%)}} &12 (3.1\%) &\textbf{\underline{910 (91.0\%)}} &90 (9.0\%) \\
    P131 (Located in the administrative territorial entity) &\textbf{\underline{264544 (57.7\%)}} &194254 (42.3\%) &\textbf{\underline{704 (79.9\%)}} &177 (20.1\%) &\textbf{\underline{552 (55.2\%)}} &448 (44.8\%) \\
    P136 (Genre) &16396 (17.0\%) &\textbf{\underline{80156 (83.0\%)}} &\textbf{\underline{547 (58.8\%)}} &384 (41.2\%) &183 (18.3\%) & \textbf{\underline{817 (81.7\%)}} \\
    P140 (Religion) &3344 (45.5\%) &\textbf{\underline{4000 (54.5\%)}} &70 (14.8\%) & \textbf{\underline{403 (85.2\%)}} &480 (48.0\%) & \textbf{\underline{520 (52.0\%)}} \\
    P159 (Headquarters location) &\textbf{\underline{24841 (69.7\%)}} &10824 (30.3\%) &\textbf{\underline{683 (70.7\%)}} &283 (29.3\%) &\textbf{\underline{685 (68.5\%)}} &315 (31.5\%) \\
    P190 (Sister city) &\textbf{\underline{2026 (54.1\%)}} &1722 (45.9\%) &\textbf{\underline{525 (52.8\%)}} &470 (47.2\%) &\textbf{\underline{542 (54.2\%)}} &458 (45.8\%) \\
    P276 (Location) &\textbf{\underline{9239 (65.5\%)}} &4867 (34.5\%) &\textbf{\underline{554 (57.9\%)}} &403 (42.1\%) &\textbf{\underline{638 (63.8\%)}} &362 (36.2\%) \\
    P413 (Position played on team / speciality) &\textbf{\underline{18307 (65.9\%)}} &9482 (34.1\%) &\textbf{\underline{751 (78.9\%)}} &201 (21.1\%) &\textbf{\underline{667 (66.7\%)}} &333 (33.3\%) \\
    P463 (Member of) &\textbf{\underline{13832 (80.0\%)}} &3452 (20.0\%) &112 (49.8\%) & \textbf{\underline{113 (50.2\%)}} &\textbf{\underline{807 (80.7\%)}} &193 (19.3\%) \\
    P495 (Country of origin) &\textbf{\underline{64856 (81.7\%)}} &14518 (18.3\%) &430 (47.4\%) & \textbf{\underline{478 (52.6\%)}} &\textbf{\underline{838 (83.8\%)}} &162 (16.2\%) \\
    P530 (Diplomatic relation) &\textbf{\underline{888 (63.7\%)}} &505 (36.3\%) &\textbf{\underline{645 (64.8\%)}} &351 (35.2\%) &\textbf{\underline{633 (63.3\%)}} &367 (36.7\%) \\
    P740 (Location of formation) &\textbf{\underline{7844 (82.5\%)}} &1663 (17.5\%) &\textbf{\underline{782 (83.7\%)}} &152 (16.3\%) &\textbf{\underline{836 (83.6\%)}} &164 (16.4\%) \\
    P937 (Work location) &\textbf{\underline{6018 (79.7\%)}} &1535 (20.3\%) &\textbf{\underline{813 (85.2\%)}} &141 (14.8\%) &\textbf{\underline{801 (80.1\%)}} &199 (19.9\%) \\
    P1001 (Applies to jurisdiction) &\textbf{\underline{2397 (60.9\%)}} &1542 (39.1\%) &\textbf{\underline{436 (62.2\%)}} &265 (37.8\%) &\textbf{\underline{612 (61.2\%)}} &388 (38.8\%) \\
    P1376 (Capital of) &1342 (30.4\%) &\textbf{\underline{3078 (69.6\%)}} &79 (33.8\%) & \textbf{\underline{155 (66.2\%)}} &297 (29.7\%) & \textbf{\underline{703 (70.3\%)}} \\
    P1412 (Languages spoken or published) &\textbf{\underline{42318 (71.2\%)}} &17137 (28.8\%) &\textbf{\underline{722 (74.5\%)}} &247 (25.5\%) &\textbf{\underline{728 (72.8\%)}} &272 (27.2\%) \\
    &&&&&&\\
    \midrule
    \textbf{Total} & \textbf{\underline{1555601 (57.1\%)}} &1168235 (42.9\%) & \textbf{\underline{13597 (63.6\%)}} &7794 (36.4\%) & \textbf{\underline{16314 (62.7\%)}} &9686 (37.3\%) \\
    &&&&&&\\
    \bottomrule
    \end{tabular}
    }
    \caption{The number and percentage of triples belonging to one of the 21 Western countries or to other countries in the T-REx, LAMA, and X-FACTR benchmarks.}
    \label{tab:representation-of-cultures-in-benchmarks}
\end{sidewaystable*}
Table \ref{tab:representation-of-cultures-in-benchmarks} provides the fine-grained percentages for the distribution of the triples of T-REx, LAMA, and X-FACTR for 21 Western countries as compared to the rest of the world. For most of the relation predicates, triples related to one of the 21 Western countries represent more than 50\% of the total triples. We find that this skewness is even larger for LAMA, and X-FACTR than for T-REx. Triples within LAMA are restricted to the ones whose objects are tokenized into a single subword by monolingual language models. This filtering might be responsible for the increased skewness of LAMA toward facts from Western countries.

\section{Augmenting the correct objects within DLAMA}
\label{sec:aug_step}
For each relation predicate, a graph is used to model all the subclass-superclass relations between the objects of the queried triples. The edges within the graph are built using Wikidata's \textbf{P279 (subclass of)} predicate. All the possible subclass/superclass relations between the list of objects for each relation predicate are queried and then used to form the edges of the graph. Afterward, the list of objects for each subject is augmented by the list of all the possible ancestors (superclasses) of these objects (e.g., The official languages of \texttt{The United States of America} are now set to \texttt{American English} and \texttt{English} instead of just \texttt{American English}).

Similarly, we noticed that the level of specificity of places of birth (objects of P19) and places of death (objects of P20) varies between different tuples. Thus, we queried all the territorial entities in which the places of birth and death are located. For instance, \texttt{Paris Hilton} had the place of birth set to \texttt{\{New York City\}} while \texttt{Donald Trump} had the place of birth set to \texttt{\{Jamaica Hospital Medical Center\}}. After querying the higher administrative-territorial entities, the set of valid objects for both entities became \texttt{\{New York City, New York, United States of America\}} and \texttt{\{Jamaica Hospital Medical Center, Queens, New York City, New York, United States of America\}} respectively.

\section{Results on Raw Triples before the Last Optional Step}
To demonstrate the impact of the last optional step within DLAMA, we evaluate the PLMs on the triples before augmenting their objects with valid overlapping ones (i.e.: before applying the optional Step \#5 of the framework). It is clear that the performance of the models shown in Table \ref{tab:overall_performance_DLAMA_raw} is worse than their performance on the augmented benchmark previously listed in Table \ref{tab:overall_performance}.

\begin{table}[t]
    \begin{subtable}{\linewidth}
    \resizebox{\columnwidth}{!}{%
        \begin{tabular}{ccccc}
        \textbf{Language} & \textbf{Model name} & $\boldsymbol{P_{@1}}$ & $\boldsymbol{P_{@1}}$ & $\boldsymbol{P_{@1}}$ \\
        \textbf{of Prompt} & & $\boldsymbol{_{Arab}}$ & $\boldsymbol{_{West}}$ & $\boldsymbol{_{All}}$\\
        &&$\boldsymbol{_{N=10946}}$&$\boldsymbol{_{N=13589}}$&$\boldsymbol{_{N=24535}}$\\
       \midrule
       \multirow{2}{*}{Arabic} & mBERT-base &    11.4 &    \textbf{12.8} &          12.2 \\
       & arBERT &    \textbf{26.6} &    19.3 &          22.6 \\
       \midrule
       \multirow{2}{*}{English} & mBERT-base &    19.1 &    \textbf{34.2} &          27.5 \\
       & BERT-base &    24.5 &   \textbf{29.9} &          27.5 \\
        \bottomrule
    \end{tabular}
    }
    \caption{DLAMA-v1 (Arab-West)}
    \end{subtable}

    \begin{subtable}{\linewidth}
    \resizebox{\columnwidth}{!}{%
    \begin{tabular}{ccccc}
    \textbf{Language} & \textbf{Model name} & $\boldsymbol{P_{@1}}$ & $\boldsymbol{P_{@1}}$ & $\boldsymbol{P_{@1}}$ \\
    \textbf{of Prompt} & & $\boldsymbol{_{Asia}}$ & $\boldsymbol{_{West}}$ & $\boldsymbol{_{All}}$\\
    &&$\boldsymbol{_{N=13479}}$&$\boldsymbol{_{N=13588}}$&$\boldsymbol{_{N=27067}}$\\
   \midrule
   \multirow{2}{*}{Korean} &  mBERT-base &      15.0 &      \textbf{22.6} &            18.8 \\
   & KyKim &      \textbf{16.0} &      11.8 &            13.9 \\
   \midrule
   \multirow{2}{*}{English} & mBERT-base &      27.1 &      \textbf{36.2} &            31.7 \\
   & BERT-base &      \textbf{36.4} &      30.4 &            33.4 \\
    \bottomrule
    \end{tabular}
    }
    \caption{DLAMA-v1 (Asia-West)}
    \end{subtable}

    \begin{subtable}{\linewidth}
    \resizebox{\columnwidth}{!}{%
    \begin{tabular}{ccccc}
    \textbf{Language} & \textbf{Model name} & $\boldsymbol{P_{@1}}$ & $\boldsymbol{P_{@1}}$ & $\boldsymbol{P_{@1}}$ \\
    \textbf{of Prompt} & & $\boldsymbol{_{S. America}}$ & $\boldsymbol{_{West}}$ & $\boldsymbol{_{All}}$\\
    &&$\boldsymbol{_{N=13071}}$&$\boldsymbol{_{N=13586}}$&$\boldsymbol{_{N=26657}}$\\
   \midrule
   \multirow{2}{*}{Spanish} &  mBERT-base &               22.3 &      \textbf{30.4} &            26.4 \\
    & BETO &               15.5 &      \textbf{25.5} &            20.6 \\
   \midrule
   \multirow{2}{*}{English} & mBERT-base &               24.1 &      \textbf{34.7} &            29.5 \\
    & BERT-base &               24.4 &      \textbf{29.9} &            27.2 \\
    \bottomrule
    \end{tabular}
    }
    \caption{DLAMA-v1 (South America-West)}
    \end{subtable}

    \caption{Performance of mBERT, and monolingual Arabic (arBERT), Korean (KyKim), Spanish (BETO), and English (BERT-base) language models on the three sets of facts of DLAMA-v1 without augmenting the set of objects (i.e.: without applying Step \#5).}
    \label{tab:overall_performance_DLAMA_raw}
\end{table}

\section{GPT3.5 performance on a subset of DLAMA (Arab-West)}
As mentioned in \S\ref{sec:gpt3.5}, we used OpenAI's API to evaluate the performance of the GPT3.5-turbo model on six predicates of DLAMA-v1 (Arab-West). The accuracy scores of the model for these predicates are reported in Table \ref{tab:GPT3.5}. We plan to extend our evaluation to cover more predicates and include other LLMs.

\begin{table}[bthp]
\centering
\scriptsize
\aboverulesep=0ex 
\belowrulesep=0ex 
\begin{tabular}{l|cc|ll|llll}
\rule{0pt}{1.1EM}
\multirow{3}{*}{\textbf{Relation}}& \multicolumn{2}{c}{$\mathbf{\#\ facts\ (entropy)}$} & \multicolumn{2}{c}{\textbf{Arabic prompts}} & \multicolumn{2}{c}{\textbf{English prompts}}\\
& & & \multicolumn{2}{c}{$Accuracy$} & \multicolumn{2}{c}{$Accuracy$}\\
& $\mathbf{_{Arab}}$ & $\mathbf{_{West}}$ & $\mathbf{_{Arab}}$ & $\mathbf{_{West}}$ & $\mathbf{_{Arab}}$ & $\mathbf{_{West}}$\\
\midrule
    \rule{0pt}{1.1EM}
    P30 (Continent) & 22 (1.0) & 19 (1.0) & 63.6 & \textbf{89.5}* & \textbf{100.0}* & 89.5 \\
    P36 (Capital) & 22 (4.5) & 19 (4.2) & \textbf{81.8}* & 63.2 & \textbf{95.5}* & 94.7 \\
    P37 (Official language) & 22 (0.0) & 19 (2.5) & \textbf{100.0}* & 89.5 & \textbf{100.0}* & \textbf{100.0}*\\
    P47 (Shares border with) & 22 (2.5) & 19 (2.7) &  \textbf{100.0}* & \textbf{100.0}*  & \textbf{95.5}* & 89.5 \\
    P190 (Sister city) & 67 (4.9) & 468 (7.3) & \textbf{6.0}* & 5.6 & 3.0 & \textbf{33.1}*  \\
    P530 (Diplomatic relation) & 22 (0.0) & 19 (0.0) & 63.6 & \textbf{68.4}* & 50.0 & \textbf{84.2}* \\
    P1376 (Capital of) & 24 (4.3) & 26 (4.0) & 87.5 & \textbf{88.5}* & \textbf{100.0}* & 92.3 \\
\bottomrule
\end{tabular}
%
\caption{The accuracy of the GPT3.5-turbo model for some predicates of the DLAMA-v1 (Arab-West) set.}
\label{tab:GPT3.5}
\end{table}

\section{Model diagnostics using the (Asia-West) and (South America-West) sets}
\label{sec:qualitative-analysis-other-models}
\noindent\textbf{Contrasting KyKim BERT to English BERT-base}: We replicate the analysis process done in \S\ref{sec:language_bias} to investigate the behavior of KyKim BERT-base and the English BERT-base models using Tables \ref{table:korean_detailed_results} and \ref{tab:language_bias_kykim_(ASIA_WEST)}. We find that the English BERT-base has the same patterns detailed before for P17, P103, P364, and P495. Moreover, since English BERT-base overgenerates \textit{Japan} for the P495 predicate, its performance on the Asian part of DLAMA-v1 (Asia-West) is high. This once again shows the importance of having two contrasting sets of facts from the same predicates. Despite the fact that the majority of triples of P495 within the Asian part of DLAMA-v1 (Asia-West) has \textit{Japan} as one of the correct labels, a biased model toward predicting \textit{Japan} has a significantly low performance on the opposing set of facts. Consequently, the bias can still be detected. 

Regarding the KyKim BERT-base model, language bias toward overpredicting \textit{Korean} is clear for the P103 and the P364 relation predications. The model also shows a bias toward the \textit{Javanese} label for P1412. This bias can be seen in the model's poor performance on the Western part of the benchmark.
P19 is a relation predicate on which the model is generally performing well. The most frequent predictions indicate that the model leans toward selecting \textit{Japan} and \textit{United States of America}. However, the model's predictions change according to the underlying culture of the triples and hence demonstrate an ability to memorize facts from both cultures.

\begin{table*}[!ptbh]
\scriptsize
\centering
\begin{tabular}{lccllll}
\multirow{3}{*}{\textbf{Relation}}&&&\multicolumn{2}{c}{\textbf{Korean prompts}} & \multicolumn{2}{c}{\textbf{English prompts}}\\
& \multicolumn{2}{c}{$\mathbf{\#\ facts\ (entropy)}$} & \multicolumn{2}{c}{$\mathbf{P_{@1}}$} & \multicolumn{2}{c}{$\mathbf{P_{@1}}$} \\
& $\mathbf{_{Asia}}$ & $\mathbf{_{West}}$ & $\mathbf{_{Asia}}$ & $\mathbf{_{West}}$ & $\mathbf{_{Asia}}$ & $\mathbf{_{West}}$\\

\midrule
                        P17 (Country) &  1000 (2.2) &  1000 (2.8) &                      37.8 & \textbf{42.1} &  \textbf{67.1} &                       45.3 \\
                 P19 (Place of birth) &  1000 (1.7) &  1000 (2.7) & \textbf{63.1} &                      55.8 &  \textbf{24.3} &                       11.9 \\
                 P20 (Place of death) &  1000 (2.6) &  1000 (2.8) &                      23.0 & \textbf{45.8} &  \textbf{40.4} &                       20.7 \\
         P27 (Country of citizenship) &  1000 (1.5) &  1000 (2.4) & \textbf{74.0} &                      53.5 &  \textbf{71.8} &                       19.5 \\
                      P30 (Continent) &    13 (0.0) &    19 (1.0) & \textbf{76.9} &                      31.6 & \textbf{100.0} &                       84.2 \\
                        P36 (Capital) &    13 (3.7) &    19 (4.2) & \textbf{30.8} &                      21.1 &                       69.2 &  \textbf{84.2} \\
              P37 (Official language) &    13 (2.7) &    19 (2.5) & \textbf{30.8} &                      26.3 &                       84.6 & \textbf{100.0} \\
             P47 (Shares border with) &    13 (1.7) &    19 (2.7) &  \textbf{0.0} &                       0.0 &                       76.9 &  \textbf{78.9} \\
               P103 (Native language) &  1000 (1.6) &  1000 (1.7) & \textbf{33.3} &                       2.3 &  \textbf{84.7} &                       75.6 \\
                    P106 (Occupation) &  1000 (0.9) &  1000 (1.0) & \textbf{17.0} &                       9.4 &                        1.4 &  \textbf{15.9} \\
                         P136 (Genre) &  1000 (1.0) &  1000 (2.5) &                       0.2 &  \textbf{0.5} &                        0.8 &   \textbf{6.3} \\
                   P190 (Sister city) &   387 (7.4) &   467 (7.3) &                       0.0 &  \textbf{1.9} &                        0.3 &   \textbf{2.8} \\
                  P264 (Record label) &  1000 (5.3) &  1000 (4.8) &  \textbf{0.3} &                       0.1 &                        3.3 &   \textbf{6.6} \\
     P364 (Original language of work) &  1000 (0.7) &  1000 (0.3) &                      10.5 & \textbf{18.5} &                       37.7 &  \textbf{89.1} \\
              P449 (Original network) &  1000 (4.6) &  1000 (5.0) &  \textbf{5.1} &                       0.2 &                        1.1 &  \textbf{10.7} \\
             P495 (Country of origin) &  1000 (0.5) &  1000 (1.3) & \textbf{29.1} &                      19.2 &  \textbf{79.7} &                        4.3 \\
           P530 (Diplomatic relation) &    13 (0.0) &    19 (0.0) &  \textbf{7.7} &                       5.3 &                       46.2 &  \textbf{68.4} \\
                   P1303 (Instrument) &  1000 (0.5) &  1000 (1.1) &                       0.4 &  \textbf{1.1} &                        9.0 &  \textbf{29.5} \\
                   P1376 (Capital of) &    27 (3.0) &    26 (4.0) & \textbf{51.9} &                      26.9 &  \textbf{88.9} &                       76.9 \\
P1412 (Languages spoken or published) &  1000 (1.3) &  1000 (1.4) &                       1.0 & \textbf{13.4} &  \textbf{87.4} &                       86.8 \\
\midrule
\textbf{Aggregated statistics} & 13479 (2.1) & 13588 (2.6) & \textbf{22.1} &                      19.5 &  \textbf{38.3} &                       31.9 \\
\bottomrule
\end{tabular}
\caption{Detailed P@1 scores of KyKim (Korean prompts) and cased BERT-base (English prompts) on the DLAMA-v1 (Asia-West) set.}
\label{table:korean_detailed_results}
\end{table*}

\begin{table*}[tbhp]
\scriptsize
\centering
\begin{tabular}{lcccccc}
\multirow{3}{*}{\textbf{Relation}}&&&\multicolumn{2}{c}{\textbf{Spanish prompts}} & \multicolumn{2}{c}{\textbf{English prompts}}\\
& \multicolumn{2}{c}{$\mathbf{\#\ facts\ (entropy)}$} & \multicolumn{2}{c}{$\mathbf{P_{@1}}$} & \multicolumn{2}{c}{$\mathbf{P_{@1}}$} \\
& $\mathbf{_{S. America}}$ & $\mathbf{_{West}}$ & $\mathbf{_{S. America}}$ & $\mathbf{_{West}}$ & $\mathbf{_{S. America}}$ & $\mathbf{_{West}}$\\

\midrule
                        P17 (Country) &  1000 (2.8) &  1000 (2.9) & \textbf{57.5} &                      47.7 &  \textbf{63.0} &                       49.9 \\
                 P19 (Place of birth) &  1000 (2.6) &  1000 (2.5) &  \textbf{2.0} &                       0.9 &  \textbf{14.6} &                        8.3 \\
                 P20 (Place of death) &  1000 (2.8) &  1000 (2.4) &                       0.1 &  \textbf{0.6} &                        0.5 &  \textbf{10.3} \\
         P27 (Country of citizenship) &  1000 (2.5) &  1000 (2.4) & \textbf{19.5} &                       4.2 &  \textbf{28.9} &                       14.5 \\
                      P30 (Continent) &    12 (0.0) &    19 (1.0) & \textbf{91.7} &                      73.7 & \textbf{100.0} &                       73.7 \\
                        P36 (Capital) &    12 (3.6) &    19 (4.2) & \textbf{83.3} &                      68.4 &                       66.7 &  \textbf{84.2} \\
              P37 (Official language) &    12 (1.2) &    19 (2.5) &                      75.0 & \textbf{84.2} &                       75.0 & \textbf{100.0} \\
             P47 (Shares border with) &    12 (1.0) &    19 (2.7) & \textbf{83.3} &                      68.4 &  \textbf{91.7} &                       78.9 \\
               P103 (Native language) &  1000 (1.1) &  1000 (1.8) &                      34.4 & \textbf{78.6} &                       58.5 &  \textbf{74.5} \\
                    P106 (Occupation) &  1000 (2.1) &  1000 (2.5) &                       6.8 &  \textbf{7.8} &                        8.3 &  \textbf{12.0} \\
                         P136 (Genre) &  1000 (2.6) &  1000 (2.4) &                       0.3 &  \textbf{1.7} &                        2.4 &   \textbf{5.5} \\
                   P190 (Sister city) &   144 (6.1) &   465 (7.4) &  \textbf{4.9} &                       1.7 &   \textbf{3.5} &                        3.0 \\
                  P264 (Record label) &   854 (6.1) &  1000 (6.0) &                       0.0 &  \textbf{0.1} &                        1.5 &   \textbf{5.6} \\
     P364 (Original language of work) &  1000 (1.1) &  1000 (0.6) &                      48.5 & \textbf{85.1} &                       60.5 &  \textbf{89.5} \\
              P449 (Original network) &  1000 (4.6) &  1000 (4.7) &                       0.3 &  \textbf{0.7} &                        0.4 &  \textbf{18.7} \\
             P495 (Country of origin) &  1000 (2.4) &  1000 (1.8) &                       6.3 & \textbf{60.0} &  \textbf{27.3} &                       10.3 \\
           P530 (Diplomatic relation) &    12 (0.0) &    19 (0.0) &                      66.7 & \textbf{68.4} &                       58.3 &  \textbf{68.4} \\
                   P1303 (Instrument) &  1000 (1.2) &  1000 (1.3) &                       6.7 & \textbf{11.7} &                       17.0 &  \textbf{26.4} \\
                   P1376 (Capital of) &    13 (3.4) &    26 (4.0) & \textbf{84.6} &                      73.1 &  \textbf{84.6} &                       76.9 \\
P1412 (Languages spoken or published) &  1000 (1.2) &  1000 (1.7) &                      20.2 & \textbf{51.6} &                       62.9 &  \textbf{89.2} \\
\midrule
\textbf{Aggregated statistics} & 13071 (2.4) & 13586 (2.7) &                      16.0 & \textbf{26.5} &                       26.9 &  \textbf{31.3} \\
\bottomrule
\end{tabular}
\caption{Detailed P@1 scores of cased BETO (Spanish prompts) and cased BERT-base (English prompts) on the DLAMA-v1 (South America-West) set.}
\label{table:south_america_detailed_results}
\end{table*}

\begin{table*}[tbp]
    \resizebox{\textwidth}{!}{%
    \centering
    \tiny
    \begin{tabular}{lcc}
        \multirow{2}{*}{\textbf{Relation predicate}} & \textbf{Common correct predictions} & \textbf{Common wrong predictions}\\
        & \textbf{(\% of predictions)} & \textbf{(\% of all predictions)}\\
        \midrule
        \multicolumn{3}{c}{\textbf{Probing KyKim with \underline{Korean prompts} populated with Asian facts}}\\
        \midrule

        P17: [X] is located in [Y] . & \colorbox{pred_purple}{\textcolor{white}{Japan}} (30.3\%) \colorbox{pred_purple}{\textcolor{white}{South Korea}} (3.0\%) \colorbox{pred_purple}{\textcolor{white}{Thailand}} (0.9\%) & \colorbox{pred_purple}{\textcolor{white}{China}} (13.8\%) \colorbox{pred_dark_orange}{United States of America} (13.0\%) \colorbox{pred_blue_bell}{\textcolor{white}{Tonga}} (9.7\%) \\
        P19: [X] was born in [Y] . & \colorbox{pred_purple}{\textcolor{white}{Japan}} (60.1\%) \colorbox{pred_purple}{\textcolor{white}{South Korea}} (1.9\%) \colorbox{pred_purple}{\textcolor{white}{South Chungcheong Province}} (0.4\%) & \colorbox{pred_dark_orange}{United States of America} (12.8\%) \colorbox{pred_purple}{\textcolor{white}{South Chungcheong Province}} (4.0\%) \colorbox{pred_purple}{\textcolor{white}{South Jeolla}} (3.2\%) \\
        P20: [X] died in [Y] . & \colorbox{pred_purple}{\textcolor{white}{Japan}} (19.2\%) \colorbox{pred_purple}{\textcolor{white}{Tokyo}} (2.7\%) \colorbox{pred_purple}{\textcolor{white}{Gyeonggi Province}} (0.3\%) & \colorbox{pred_dark_orange}{United States of America} (19.8\%) \colorbox{pred_purple}{\textcolor{white}{Gyeonggi Province}} (14.5\%) \colorbox{pred_dark_orange}{Germany} (4.7\%) \\
        P27: [X] is [Y] citizen . & \colorbox{pred_purple}{\textcolor{white}{Japan}} (70.2\%) \colorbox{pred_purple}{\textcolor{white}{South Korea}} (3.2\%) \colorbox{pred_purple}{\textcolor{white}{Singapore}} (0.2\%) & \colorbox{pred_purple}{\textcolor{white}{Korea}} (13.5\%) \colorbox{pred_purple}{\textcolor{white}{South Korea}} (5.0\%) \colorbox{pred_purple}{\textcolor{white}{China}} (3.0\%) \\
        P495: [X] was created in [Y] . & \colorbox{pred_purple}{\textcolor{white}{Japan}} (26.2\%) \colorbox{pred_purple}{\textcolor{white}{South Korea}} (2.9\%) & \colorbox{pred_blue_bell}{\textcolor{white}{Jordan}} (29.7\%) \colorbox{pred_purple}{\textcolor{white}{South Korea}} (28.0\%) \colorbox{pred_dark_orange}{United States of America} (6.4\%) \\
        P103: The native language of [X] is [Y] . & \colorbox{pred_purple}{\textcolor{white}{Korean}} (31.5\%) \colorbox{pred_purple}{\textcolor{white}{Japanese}} (1.5\%) \colorbox{pred_purple}{\textcolor{white}{Chinese}} (0.2\%) & \colorbox{pred_purple}{\textcolor{white}{Korean}} (66.5\%) \colorbox{pred_purple}{\textcolor{white}{Hakka}} (0.1\%) \colorbox{pred_purple}{\textcolor{white}{Chinese}} (0.1\%) \\
        P364: The original language of [X] is [Y] . & \colorbox{pred_purple}{\textcolor{white}{Korean}} (6.0\%) \colorbox{pred_purple}{\textcolor{white}{Japanese}} (4.3\%) \colorbox{pred_purple}{\textcolor{white}{Chinese}} (0.1\%) & \colorbox{pred_purple}{\textcolor{white}{Korean}} (79.9\%) \colorbox{pred_dark_orange}{English} (8.1\%) \colorbox{pred_dark_orange}{German} (0.4\%) \\
        P1412: [X] used to communicate in [Y] . & \colorbox{pred_purple}{\textcolor{white}{Vietnamese}} (0.4\%) \colorbox{pred_purple}{\textcolor{white}{Javanese}} (0.3\%) \colorbox{pred_purple}{\textcolor{white}{Japanese}} (0.1\%) & \colorbox{pred_purple}{\textcolor{white}{Javanese}} (77.2\%) \colorbox{pred_blue_bell}{\textcolor{white}{Tamil}} (4.3\%) \colorbox{pred_purple}{\textcolor{white}{Wu Chinese}} (3.9\%) \\

        \midrule
        \multicolumn{3}{c}{\textbf{Probing KyKim with \underline{Korean prompts} populated with Western facts}}\\
        \midrule

        P17: [X] is located in [Y] . & \colorbox{pred_purple}{\textcolor{white}{United States of America}} (28.2\%) \colorbox{pred_purple}{\textcolor{white}{France}} (4.9\%) \colorbox{pred_purple}{\textcolor{white}{Germany}} (4.0\%) & \colorbox{pred_purple}{\textcolor{white}{United States of America}} (27.4\%) \colorbox{pred_dark_orange}{Korea} (7.6\%) \colorbox{pred_dark_orange}{China} (5.7\%) \\
        P19: [X] was born in [Y] . & \colorbox{pred_purple}{\textcolor{white}{United States of America}} (42.0\%) \colorbox{pred_purple}{\textcolor{white}{France}} (5.6\%) \colorbox{pred_purple}{\textcolor{white}{Italy}} (4.4\%) & \colorbox{pred_purple}{\textcolor{white}{United States of America}} (27.0\%) \colorbox{pred_purple}{\textcolor{white}{Italy}} (7.7\%) \colorbox{pred_purple}{\textcolor{white}{Germany}} (6.8\%) \\
        P20: [X] died in [Y] . & \colorbox{pred_purple}{\textcolor{white}{United States of America}} (29.2\%) \colorbox{pred_purple}{\textcolor{white}{Germany}} (7.0\%) \colorbox{pred_purple}{\textcolor{white}{France}} (5.4\%) & \colorbox{pred_purple}{\textcolor{white}{Germany}} (22.2\%) \colorbox{pred_purple}{\textcolor{white}{United States of America}} (17.7\%) \colorbox{pred_purple}{\textcolor{white}{Italy}} (3.2\%) \\
        P27: [X] is [Y] citizen . & \colorbox{pred_purple}{\textcolor{white}{United States of America}} (37.7\%) \colorbox{pred_purple}{\textcolor{white}{France}} (11.8\%) \colorbox{pred_purple}{\textcolor{white}{Italy}} (1.8\%) & \colorbox{pred_purple}{\textcolor{white}{United States of America}} (15.3\%) \colorbox{pred_purple}{\textcolor{white}{France}} (10.5\%) \colorbox{pred_dark_orange}{Korea} (9.3\%) \\
        P495: [X] was created in [Y] . & \colorbox{pred_purple}{\textcolor{white}{United States of America}} (17.9\%) \colorbox{pred_purple}{\textcolor{white}{Germany}} (0.4\%) \colorbox{pred_dark_orange}{Japan} (0.4\%) & \colorbox{pred_dark_orange}{South Korea} (36.4\%) \colorbox{pred_light_orange}{Jordan} (21.4\%) \colorbox{pred_dark_orange}{Japan} (12.8\%) \\
        P103: The native language of [X] is [Y] . & \colorbox{pred_purple}{\textcolor{white}{English}} (2.0\%) \colorbox{pred_purple}{\textcolor{white}{French}} (0.3\%) & \colorbox{pred_dark_orange}{Korean} (97.0\%) \colorbox{pred_purple}{\textcolor{white}{English}} (0.4\%) \colorbox{pred_dark_orange}{Japanese} (0.2\%) \\
        P364: The original language of [X] is [Y] . & \colorbox{pred_purple}{\textcolor{white}{English}} (18.0\%) \colorbox{pred_dark_orange}{Korean} (0.3\%) \colorbox{pred_purple}{\textcolor{white}{French}} (0.1\%) & \colorbox{pred_dark_orange}{Korean} (79.3\%) \colorbox{pred_purple}{\textcolor{white}{English}} (0.8\%) \colorbox{pred_purple}{\textcolor{white}{French}} (0.7\%) \\
        P1412: [X] used to communicate in [Y] . & \colorbox{pred_purple}{\textcolor{white}{French}} (7.4\%) \colorbox{pred_purple}{\textcolor{white}{German}} (4.9\%) \colorbox{pred_purple}{\textcolor{white}{Spanish}} (0.5\%) & \colorbox{pred_dark_orange}{Javanese} (51.2\%) \colorbox{pred_purple}{\textcolor{white}{German}} (9.3\%) \colorbox{pred_dark_orange}{Burmese} (9.0\%) \\

        \midrule
                \multirow{2}{*}{\textbf{Relation predicate}} & \textbf{Common correct predictions} & \textbf{Common wrong predictions}\\
        & \textbf{(\% of predictions)} & \textbf{(\% of all predictions)}\\
        \midrule
        \multicolumn{3}{c}{\textbf{Probing BERT-base with \underline{English prompts} populated with Asian facts}}\\
        \midrule

        P17: [X] is located in [Y] . & \colorbox{pred_purple}{\textcolor{white}{Japan}} (48.9\%) \colorbox{pred_purple}{\textcolor{white}{Thailand}} (3.4\%) \colorbox{pred_purple}{\textcolor{white}{Taiwan}} (3.2\%) & \colorbox{pred_purple}{\textcolor{white}{China}} (7.5\%) \colorbox{pred_blue_bell}{\textcolor{white}{Moscow}} (5.6\%) \colorbox{pred_purple}{\textcolor{white}{Taiwan}} (3.2\%) \\
        P19: [X] was born in [Y] . & \colorbox{pred_purple}{\textcolor{white}{Tokyo}} (17.8\%) \colorbox{pred_purple}{\textcolor{white}{Seoul}} (2.4\%) \colorbox{pred_purple}{\textcolor{white}{Vietnam}} (1.0\%) & \colorbox{pred_purple}{\textcolor{white}{Tokyo}} (52.3\%) \colorbox{pred_purple}{\textcolor{white}{Seoul}} (7.0\%) \colorbox{pred_purple}{\textcolor{white}{Beijing}} (4.1\%) \\
        P20: [X] died in [Y] . & \colorbox{pred_purple}{\textcolor{white}{Tokyo}} (26.3\%) \colorbox{pred_purple}{\textcolor{white}{Beijing}} (5.8\%) \colorbox{pred_purple}{\textcolor{white}{Seoul}} (4.8\%) & \colorbox{pred_purple}{\textcolor{white}{Beijing}} (21.0\%) \colorbox{pred_purple}{\textcolor{white}{Tokyo}} (16.6\%) \colorbox{pred_purple}{\textcolor{white}{Paris}} (7.8\%) \\
        P27: [X] is [Y] citizen . & \colorbox{pred_purple}{\textcolor{white}{Japan}} (67.4\%) \colorbox{pred_purple}{\textcolor{white}{Taiwan}} (2.4\%) \colorbox{pred_purple}{\textcolor{white}{Vietnam}} (1.0\%) & \colorbox{pred_purple}{\textcolor{white}{Taiwan}} (12.8\%) \colorbox{pred_purple}{\textcolor{white}{Singapore}} (5.4\%) \colorbox{pred_purple}{\textcolor{white}{Korea}} (4.6\%) \\
        P495: [X] was created in [Y] . & \colorbox{pred_purple}{\textcolor{white}{Japan}} (79.5\%) \colorbox{pred_purple}{\textcolor{white}{Vietnam}} (0.1\%) \colorbox{pred_purple}{\textcolor{white}{Thailand}} (0.1\%) & \colorbox{pred_purple}{\textcolor{white}{Japan}} (5.3\%) \colorbox{pred_purple}{\textcolor{white}{India}} (3.5\%) \colorbox{pred_purple}{\textcolor{white}{Germany}} (3.0\%) \\
        P103: The native language of [X] is [Y] . & \colorbox{pred_purple}{\textcolor{white}{Japanese}} (52.4\%) \colorbox{pred_purple}{\textcolor{white}{Korean}} (26.4\%) \colorbox{pred_purple}{\textcolor{white}{Chinese}} (3.8\%) & \colorbox{pred_dark_orange}{English} (4.2\%) \colorbox{pred_dark_orange}{Spanish} (1.6\%) \colorbox{pred_purple}{\textcolor{white}{Wu Chinese}} (1.5\%) \\
        P364: The original language of [X] is [Y] . & \colorbox{pred_purple}{\textcolor{white}{Japanese}} (34.6\%) \colorbox{pred_dark_orange}{English} (2.3\%) \colorbox{pred_purple}{\textcolor{white}{Chinese}} (0.3\%) & \colorbox{pred_dark_orange}{English} (50.2\%) \colorbox{pred_dark_orange}{French} (1.8\%) \colorbox{pred_dark_orange}{Latin} (1.5\%) \\
        P1412: [X] used to communicate in [Y] . & \colorbox{pred_purple}{\textcolor{white}{Japanese}} (73.5\%) \colorbox{pred_purple}{\textcolor{white}{Korean}} (6.8\%) \colorbox{pred_purple}{\textcolor{white}{Chinese}} (2.4\%) & \colorbox{pred_dark_orange}{English} (5.6\%) \colorbox{pred_purple}{\textcolor{white}{Cantonese}} (2.2\%) \colorbox{pred_purple}{\textcolor{white}{Japanese}} (1.5\%) \\

        \midrule    
        \multicolumn{3}{c}{\textbf{Probing BERT-base with \underline{English prompts} populated with Western facts}}\\
        \midrule

        P17: [X] is located in [Y] . & \colorbox{pred_purple}{\textcolor{white}{France}} (15.5\%) \colorbox{pred_purple}{\textcolor{white}{Germany}} (8.9\%) \colorbox{pred_purple}{\textcolor{white}{Spain}} (8.3\%) & \colorbox{pred_light_orange}{Georgia} (12.6\%) \colorbox{pred_blue_bell}{\textcolor{white}{Moscow}} (7.7\%) \colorbox{pred_purple}{\textcolor{white}{Canada}} (6.3\%) \\
        P19: [X] was born in [Y] . & \colorbox{pred_purple}{\textcolor{white}{Paris}} (4.5\%) \colorbox{pred_purple}{\textcolor{white}{London}} (1.2\%) \colorbox{pred_purple}{\textcolor{white}{Rome}} (1.0\%) & \colorbox{pred_purple}{\textcolor{white}{Chicago}} (24.2\%) \colorbox{pred_purple}{\textcolor{white}{London}} (21.4\%) \colorbox{pred_purple}{\textcolor{white}{Paris}} (10.5\%) \\
        P20: [X] died in [Y] . & \colorbox{pred_purple}{\textcolor{white}{Paris}} (11.2\%) \colorbox{pred_purple}{\textcolor{white}{Rome}} (3.7\%) \colorbox{pred_purple}{\textcolor{white}{London}} (2.5\%) & \colorbox{pred_purple}{\textcolor{white}{Paris}} (30.7\%) \colorbox{pred_purple}{\textcolor{white}{London}} (19.9\%) \colorbox{pred_purple}{\textcolor{white}{Rome}} (9.4\%) \\
        P27: [X] is [Y] citizen . & \colorbox{pred_purple}{\textcolor{white}{France}} (11.8\%) \colorbox{pred_purple}{\textcolor{white}{Italy}} (3.4\%) \colorbox{pred_purple}{\textcolor{white}{Canada}} (1.0\%) & \colorbox{pred_purple}{\textcolor{white}{British America}} (35.8\%) \colorbox{pred_dark_orange}{Singapore} (19.7\%) \colorbox{pred_purple}{\textcolor{white}{Austria}} (4.9\%) \\
        P495: [X] was created in [Y] . & \colorbox{pred_purple}{\textcolor{white}{France}} (1.5\%) \colorbox{pred_purple}{\textcolor{white}{Germany}} (1.0\%) \colorbox{pred_purple}{\textcolor{white}{Japan}} (0.5\%) & \colorbox{pred_dark_orange}{Japan} (60.2\%) \colorbox{pred_purple}{\textcolor{white}{England}} (10.7\%) \colorbox{pred_purple}{\textcolor{white}{Germany}} (5.0\%) \\
        P103: The native language of [X] is [Y] . & \colorbox{pred_purple}{\textcolor{white}{English}} (53.1\%) \colorbox{pred_purple}{\textcolor{white}{French}} (12.6\%) \colorbox{pred_purple}{\textcolor{white}{German}} (3.4\%) & \colorbox{pred_purple}{\textcolor{white}{Spanish}} (6.9\%) \colorbox{pred_purple}{\textcolor{white}{French}} (3.9\%) \colorbox{pred_purple}{\textcolor{white}{German}} (3.7\%) \\
        P364: The original language of [X] is [Y] . & \colorbox{pred_purple}{\textcolor{white}{English}} (87.0\%) \colorbox{pred_purple}{\textcolor{white}{French}} (0.7\%) \colorbox{pred_purple}{\textcolor{white}{German}} (0.5\%) & \colorbox{pred_purple}{\textcolor{white}{Latin}} (2.3\%) \colorbox{pred_purple}{\textcolor{white}{English}} (1.9\%) \colorbox{pred_purple}{\textcolor{white}{French}} (1.7\%) \\
        P1412: [X] used to communicate in [Y] . & \colorbox{pred_purple}{\textcolor{white}{English}} (61.7\%) \colorbox{pred_purple}{\textcolor{white}{French}} (12.1\%) \colorbox{pred_purple}{\textcolor{white}{German}} (4.3\%) & \colorbox{pred_purple}{\textcolor{white}{English}} (4.6\%) \colorbox{pred_purple}{\textcolor{white}{Spanish}} (2.4\%) \colorbox{pred_light_orange}{Arabic} (1.4\%) \\

    \bottomrule
    \end{tabular}
    } 
    \caption{The most common predictions for monolingual Korean and English BERT models when probed by DLAMA-v1 (Asia-West) with English and Korean prompts, respectively. \colorbox{pred_purple}{\textcolor{white}{Purple}} culturally related prediction, \colorbox{pred_blue_bell}{\textcolor{white}{Blue bell}} culturally proximate prediction, \colorbox{pred_light_orange}{Light Orange} culturally proximate prediction to another culture, \colorbox{pred_dark_orange}{Orange} culturally related prediction to the other culture. \textbf{Note:} The Korean prompts/entities are translated for clarity.}
    \label{tab:language_bias_kykim_(ASIA_WEST)}
\end{table*}

\noindent\textbf{Contrasting Spanish BETO to English BERT-base}: 
While similar patterns can be found in  Tables \ref{table:south_america_detailed_results}, and \ref{tab:language_bias_beto_(SOUTH_AMERICA_WEST)}, a new subtle bias is that BERT-base predicts \texttt{Madrid} for more than 50\% of the South American triples in P19 (Place of Birth), and P20 (Place of Death) predicates. This might be attributed to the fact that South American names are hard to distinguish from Spanish ones.

\begin{table*}[tbhp]
    \resizebox{\textwidth}{!}{%
    \centering
    \tiny
    \begin{tabular}{lcc}
            \multirow{2}{*}{\textbf{Relation predicate}} & \textbf{Common correct predictions} & \textbf{Common wrong predictions}\\
        & \textbf{(\% of predictions)} & \textbf{(\% of all predictions)}\\
        \midrule
        \multicolumn{3}{c}{\textbf{Probing BETO with \underline{Spanish prompts} populated with South America facts}}\\
        \midrule

        P17: [X] is located in [Y] . & \colorbox{pred_purple}{\textcolor{white}{Brazil}} (17.6\%) \colorbox{pred_purple}{\textcolor{white}{Argentina}} (14.9\%) \colorbox{pred_purple}{\textcolor{white}{Chile}} (7.9\%) & \colorbox{pred_blue_bell}{\textcolor{white}{Mexico}} (12.0\%) \colorbox{pred_blue_bell}{\textcolor{white}{Curaçao}} (8.1\%) \colorbox{pred_purple}{\textcolor{white}{Venezuela}} (4.3\%) \\
        P19: [X] was born in [Y] . & \colorbox{pred_purple}{\textcolor{white}{Buenos Aires}} (1.5\%) \colorbox{pred_purple}{\textcolor{white}{Lima}} (0.2\%) \colorbox{pred_purple}{\textcolor{white}{Brazil}} (0.1\%) & \colorbox{pred_dark_orange}{Altötting} (91.0\%) \colorbox{pred_purple}{\textcolor{white}{Buenos Aires}} (5.6\%) \colorbox{pred_dark_orange}{Madrid} (0.3\%) \\
        P20: [X] died in [Y] . & \colorbox{pred_purple}{\textcolor{white}{Aripuanã}} (0.1\%) & \colorbox{pred_purple}{\textcolor{white}{Aripuanã}} (99.6\%) \colorbox{pred_purple}{\textcolor{white}{Buenos Aires}} (0.1\%) \colorbox{pred_purple}{\textcolor{white}{Caracas}} (0.1\%) \\
        P27: [X] is [Y] citizen . & \colorbox{pred_purple}{\textcolor{white}{Brazil}} (13.0\%) \colorbox{pred_purple}{\textcolor{white}{Colombia}} (4.3\%) \colorbox{pred_purple}{\textcolor{white}{Chile}} (1.4\%) & \colorbox{pred_purple}{\textcolor{white}{Colombia}} (39.7\%) \colorbox{pred_light_orange}{Taiwan} (9.0\%) \colorbox{pred_blue_bell}{\textcolor{white}{Mexico}} (5.5\%) \\
        P495: [X] was created in [Y] . & \colorbox{pred_purple}{\textcolor{white}{Argentina}} (1.3\%) \colorbox{pred_purple}{\textcolor{white}{Chile}} (1.2\%) \colorbox{pred_purple}{\textcolor{white}{Brazil}} (1.1\%) & \colorbox{pred_dark_orange}{United States of America} (29.6\%) \colorbox{pred_purple}{\textcolor{white}{Río de la Plata}} (23.4\%) \colorbox{pred_dark_orange}{Kingdom of Portugal} (16.6\%) \\
        P103: The native language of [X] is [Y] . & \colorbox{pred_purple}{\textcolor{white}{Spanish}} (17.6\%) \colorbox{pred_purple}{\textcolor{white}{Portuguese}} (15.4\%) \colorbox{pred_dark_orange}{English} (1.4\%) & \colorbox{pred_dark_orange}{English} (48.2\%) \colorbox{pred_purple}{\textcolor{white}{Spanish}} (14.1\%) \colorbox{pred_dark_orange}{French} (2.2\%) \\
        P364: The original language of [X] is [Y] . & \colorbox{pred_purple}{\textcolor{white}{Spanish}} (39.1\%) \colorbox{pred_purple}{\textcolor{white}{Portuguese}} (7.5\%) \colorbox{pred_dark_orange}{English} (1.9\%) & \colorbox{pred_dark_orange}{English} (36.1\%) \colorbox{pred_purple}{\textcolor{white}{Spanish}} (13.7\%) \colorbox{pred_dark_orange}{French} (0.5\%) \\
        P1412: [X] used to communicate in [Y] . & \colorbox{pred_purple}{\textcolor{white}{Spanish}} (13.4\%) \colorbox{pred_dark_orange}{English} (6.5\%) \colorbox{pred_purple}{\textcolor{white}{Portuguese}} (0.2\%) & \colorbox{pred_dark_orange}{English} (70.6\%) \colorbox{pred_purple}{\textcolor{white}{Spanish}} (5.8\%) \colorbox{pred_dark_orange}{Latin} (2.6\%) \\
        
        \midrule
        \multicolumn{3}{c}{\textbf{Probing BETO with \underline{Spanish prompts} populated with Western facts}}\\
        \midrule

        P17: [X] is located in [Y] . & \colorbox{pred_purple}{\textcolor{white}{France}} (13.1\%) \colorbox{pred_purple}{\textcolor{white}{Spain}} (10.6\%) \colorbox{pred_purple}{\textcolor{white}{United States of America}} (10.2\%) & \colorbox{pred_light_orange}{Mexico} (15.5\%) \colorbox{pred_purple}{\textcolor{white}{United States of America}} (5.7\%) \colorbox{pred_purple}{\textcolor{white}{Canada}} (3.2\%) \\
        P19: [X] was born in [Y] . & \colorbox{pred_purple}{\textcolor{white}{Paris}} (0.5\%) \colorbox{pred_purple}{\textcolor{white}{Rome}} (0.2\%) \colorbox{pred_purple}{\textcolor{white}{Altötting}} (0.1\%) & \colorbox{pred_purple}{\textcolor{white}{Altötting}} (95.6\%) \colorbox{pred_purple}{\textcolor{white}{Paris}} (2.4\%) \colorbox{pred_purple}{\textcolor{white}{Rome}} (0.4\%) \\
        P20: [X] died in [Y] . & \colorbox{pred_purple}{\textcolor{white}{Paris}} (0.3\%) \colorbox{pred_purple}{\textcolor{white}{Rome}} (0.2\%) \colorbox{pred_purple}{\textcolor{white}{Madrid}} (0.1\%) & \colorbox{pred_dark_orange}{Aripuanã} (98.8\%) \colorbox{pred_purple}{\textcolor{white}{Paris}} (0.2\%) \colorbox{pred_purple}{\textcolor{white}{Rome}} (0.1\%) \\
        P27: [X] is [Y] citizen . & \colorbox{pred_purple}{\textcolor{white}{France}} (1.4\%) \colorbox{pred_purple}{\textcolor{white}{Italy}} (1.1\%) \colorbox{pred_purple}{\textcolor{white}{Spain}} (0.5\%) & \colorbox{pred_light_orange}{Taiwan} (25.3\%) \colorbox{pred_purple}{\textcolor{white}{Australia}} (21.7\%) \colorbox{pred_blue_bell}{\textcolor{white}{Socialist Republic of Romania}} (7.3\%) \\
        P495: [X] was created in [Y] . & \colorbox{pred_purple}{\textcolor{white}{United States of America}} (54.6\%) \colorbox{pred_purple}{\textcolor{white}{France}} (2.6\%) \colorbox{pred_purple}{\textcolor{white}{Spain}} (2.3\%) & \colorbox{pred_purple}{\textcolor{white}{United States of America}} (12.5\%) \colorbox{pred_purple}{\textcolor{white}{Kingdom of Portugal}} (5.1\%) \colorbox{pred_dark_orange}{Río de la Plata} (4.9\%) \\
        P103: The native language of [X] is [Y] . & \colorbox{pred_purple}{\textcolor{white}{English}} (63.3\%) \colorbox{pred_purple}{\textcolor{white}{French}} (10.7\%) \colorbox{pred_purple}{\textcolor{white}{German}} (1.7\%) & \colorbox{pred_purple}{\textcolor{white}{English}} (18.4\%) \colorbox{pred_purple}{\textcolor{white}{French}} (1.2\%) \colorbox{pred_purple}{\textcolor{white}{Spanish}} (1.2\%) \\
        P364: The original language of [X] is [Y] . & \colorbox{pred_purple}{\textcolor{white}{English}} (80.4\%) \colorbox{pred_purple}{\textcolor{white}{Spanish}} (2.8\%) \colorbox{pred_purple}{\textcolor{white}{Italian}} (0.7\%) & \colorbox{pred_purple}{\textcolor{white}{Spanish}} (10.5\%) \colorbox{pred_purple}{\textcolor{white}{English}} (3.5\%) \colorbox{pred_purple}{\textcolor{white}{French}} (0.3\%) \\
        P1412: [X] used to communicate in [Y] . & \colorbox{pred_purple}{\textcolor{white}{English}} (41.3\%) \colorbox{pred_purple}{\textcolor{white}{French}} (6.0\%) \colorbox{pred_purple}{\textcolor{white}{German}} (2.9\%) & \colorbox{pred_purple}{\textcolor{white}{English}} (43.6\%) \colorbox{pred_purple}{\textcolor{white}{Spanish}} (3.1\%) \colorbox{pred_purple}{\textcolor{white}{Latin}} (1.3\%) \\        

        \midrule
        \multirow{2}{*}{\textbf{Relation predicate}} & \textbf{Common correct predictions} & \textbf{Common wrong predictions}\\
        & \textbf{(\% of predictions)} & \textbf{(\% of all predictions)}\\
        \midrule
        \multicolumn{3}{c}{\textbf{Probing BERT-base with \underline{English prompts} populated with South American facts}}\\
        \midrule

        P17: [X] is located in [Y] . & \colorbox{pred_purple}{\textcolor{white}{Brazil}} (18.1\%) \colorbox{pred_purple}{\textcolor{white}{Argentina}} (16.3\%) \colorbox{pred_purple}{\textcolor{white}{Chile}} (8.1\%) & \colorbox{pred_purple}{\textcolor{white}{Bolivia}} (6.5\%) \colorbox{pred_blue_bell}{\textcolor{white}{Mexico}} (5.1\%) \colorbox{pred_dark_orange}{Spain} (3.5\%) \\
        P19: [X] was born in [Y] . & \colorbox{pred_purple}{\textcolor{white}{Brazil}} (14.0\%) \colorbox{pred_purple}{\textcolor{white}{Argentina}} (0.3\%) \colorbox{pred_purple}{\textcolor{white}{Bolivia}} (0.1\%) & \colorbox{pred_dark_orange}{Madrid} (54.1\%) \colorbox{pred_dark_orange}{Rome} (6.4\%) \colorbox{pred_dark_orange}{Milan} (3.9\%) \\
        P20: [X] died in [Y] . & \colorbox{pred_purple}{\textcolor{white}{Peru}} (0.2\%) \colorbox{pred_purple}{\textcolor{white}{Brazil}} (0.2\%) \colorbox{pred_purple}{\textcolor{white}{London}} (0.1\%) & \colorbox{pred_dark_orange}{Madrid} (55.1\%) \colorbox{pred_dark_orange}{Paris} (16.8\%) \colorbox{pred_dark_orange}{Rome} (11.4\%) \\
        P27: [X] is [Y] citizen . & \colorbox{pred_purple}{\textcolor{white}{Brazil}} (18.0\%) \colorbox{pred_purple}{\textcolor{white}{Argentina}} (9.9\%) \colorbox{pred_purple}{\textcolor{white}{Italy}} (0.3\%) & \colorbox{pred_blue_bell}{\textcolor{white}{Mexico}} (25.0\%) \colorbox{pred_purple}{\textcolor{white}{Argentina}} (15.8\%) \colorbox{pred_blue_bell}{\textcolor{white}{Honduras}} (6.0\%) \\
        P495: [X] was created in [Y] . & \colorbox{pred_purple}{\textcolor{white}{Brazil}} (19.7\%) \colorbox{pred_purple}{\textcolor{white}{Argentina}} (2.2\%) \colorbox{pred_purple}{\textcolor{white}{Chile}} (1.8\%) & \colorbox{pred_blue_bell}{\textcolor{white}{Mexico}} (24.4\%) \colorbox{pred_light_orange}{Japan} (11.3\%) \colorbox{pred_dark_orange}{Spain} (8.7\%) \\
        P103: The native language of [X] is [Y] . & \colorbox{pred_purple}{\textcolor{white}{Portuguese}} (34.2\%) \colorbox{pred_purple}{\textcolor{white}{Spanish}} (23.2\%) \colorbox{pred_dark_orange}{English} (0.6\%) & \colorbox{pred_purple}{\textcolor{white}{Spanish}} (20.5\%) \colorbox{pred_dark_orange}{Italian} (4.8\%) \colorbox{pred_dark_orange}{English} (3.6\%) \\
        P364: The original language of [X] is [Y] . & \colorbox{pred_purple}{\textcolor{white}{Spanish}} (40.7\%) \colorbox{pred_purple}{\textcolor{white}{Portuguese}} (17.1\%) \colorbox{pred_dark_orange}{English} (2.6\%) & \colorbox{pred_dark_orange}{English} (19.4\%) \colorbox{pred_purple}{\textcolor{white}{Spanish}} (6.6\%) \colorbox{pred_dark_orange}{Latin} (5.5\%) \\
        P1412: [X] used to communicate in [Y] . & \colorbox{pred_purple}{\textcolor{white}{Spanish}} (44.7\%) \colorbox{pred_purple}{\textcolor{white}{Portuguese}} (14.5\%) \colorbox{pred_dark_orange}{English} (2.6\%) & \colorbox{pred_purple}{\textcolor{white}{Spanish}} (16.4\%) \colorbox{pred_dark_orange}{English} (12.0\%) \colorbox{pred_dark_orange}{Italian} (3.0\%) \\

        \midrule    
        \multicolumn{3}{c}{\textbf{Probing BERT-base with \underline{English prompts} populated with Western facts}}\\
        \midrule

        P17: [X] is located in [Y] . & \colorbox{pred_purple}{\textcolor{white}{France}} (16.9\%) \colorbox{pred_purple}{\textcolor{white}{Spain}} (12.7\%) \colorbox{pred_purple}{\textcolor{white}{Germany}} (7.9\%) & \colorbox{pred_blue_bell}{\textcolor{white}{Georgia}} (9.5\%) \colorbox{pred_purple}{\textcolor{white}{Canada}} (5.1\%) \colorbox{pred_light_orange}{Lebanon} (2.9\%) \\
        P19: [X] was born in [Y] . & \colorbox{pred_purple}{\textcolor{white}{Berlin}} (2.3\%) \colorbox{pred_purple}{\textcolor{white}{Paris}} (1.7\%) \colorbox{pred_purple}{\textcolor{white}{London}} (1.7\%) & \colorbox{pred_purple}{\textcolor{white}{Berlin}} (36.7\%) \colorbox{pred_purple}{\textcolor{white}{Chicago}} (13.0\%) \colorbox{pred_purple}{\textcolor{white}{London}} (12.9\%) \\
        P20: [X] died in [Y] . & \colorbox{pred_purple}{\textcolor{white}{Paris}} (5.1\%) \colorbox{pred_purple}{\textcolor{white}{London}} (1.8\%) \colorbox{pred_purple}{\textcolor{white}{Rome}} (1.1\%) & \colorbox{pred_purple}{\textcolor{white}{Munich}} (23.0\%) \colorbox{pred_purple}{\textcolor{white}{Paris}} (22.8\%) \colorbox{pred_purple}{\textcolor{white}{Berlin}} (15.0\%) \\
        P27: [X] is [Y] citizen . & \colorbox{pred_purple}{\textcolor{white}{France}} (8.6\%) \colorbox{pred_purple}{\textcolor{white}{Austria}} (2.0\%) \colorbox{pred_purple}{\textcolor{white}{Italy}} (1.9\%) & \colorbox{pred_purple}{\textcolor{white}{Austria}} (36.8\%) \colorbox{pred_purple}{\textcolor{white}{British America}} (26.1\%) \colorbox{pred_purple}{\textcolor{white}{Netherlands}} (4.2\%) \\
        P495: [X] was created in [Y] . & \colorbox{pred_purple}{\textcolor{white}{France}} (4.8\%) \colorbox{pred_purple}{\textcolor{white}{Spain}} (1.6\%) \colorbox{pred_purple}{\textcolor{white}{Germany}} (1.5\%) & \colorbox{pred_light_orange}{Japan} (53.1\%) \colorbox{pred_purple}{\textcolor{white}{England}} (10.3\%) \colorbox{pred_purple}{\textcolor{white}{Germany}} (5.3\%) \\
        P103: The native language of [X] is [Y] . & \colorbox{pred_purple}{\textcolor{white}{English}} (48.8\%) \colorbox{pred_purple}{\textcolor{white}{French}} (15.1\%) \colorbox{pred_purple}{\textcolor{white}{German}} (3.4\%) & \colorbox{pred_purple}{\textcolor{white}{Spanish}} (7.4\%) \colorbox{pred_purple}{\textcolor{white}{German}} (3.7\%) \colorbox{pred_purple}{\textcolor{white}{French}} (3.0\%) \\
        P364: The original language of [X] is [Y] . & \colorbox{pred_purple}{\textcolor{white}{English}} (83.8\%) \colorbox{pred_purple}{\textcolor{white}{Spanish}} (2.9\%) \colorbox{pred_purple}{\textcolor{white}{German}} (1.3\%) & \colorbox{pred_purple}{\textcolor{white}{English}} (2.3\%) \colorbox{pred_purple}{\textcolor{white}{Latin}} (2.0\%) \colorbox{pred_purple}{\textcolor{white}{French}} (1.8\%) \\
        P1412: [X] used to communicate in [Y] . & \colorbox{pred_purple}{\textcolor{white}{English}} (38.8\%) \colorbox{pred_purple}{\textcolor{white}{German}} (34.0\%) \colorbox{pred_purple}{\textcolor{white}{French}} (10.1\%) & \colorbox{pred_purple}{\textcolor{white}{English}} (5.4\%) \colorbox{pred_purple}{\textcolor{white}{Spanish}} (1.3\%) \colorbox{pred_purple}{\textcolor{white}{German}} (0.8\%) \\        

    \bottomrule
    \end{tabular}
    } 
    \caption{The most common predictions for monolingual Spanish and English BERT models when probed by DLAMA-v1 (South America-West) with English and Spanish prompts, respectively. \colorbox{pred_purple}{\textcolor{white}{Purple}} culturally related prediction, \colorbox{pred_blue_bell}{\textcolor{white}{Blue bell}} culturally proximate prediction, \colorbox{pred_light_orange}{Light Orange} culturally proximate prediction to another culture, \colorbox{pred_dark_orange}{Orange} culturally related prediction to the other culture. \textbf{Note:} The Spanish prompts/entities are translated for clarity.}
    \label{tab:language_bias_beto_(SOUTH_AMERICA_WEST)}
\end{table*}

\section{Details of DLAMA}
\noindent \textbf{Wikipedia sites}: For the Arab, Asian, South American, and Western cultures, representative countries from each region are used as a proxy. Table \ref{tab:wikipedia_articles_languages} enumerates the countries representing these cultures and their relevant respective Wikipedia sites.

\noindent \textbf{Probing templates}: To probe the models' factual knowledge, natural language templates are used to transform the triples into prompts. The template has two fields for the subject $[X]$ and the object $[Y]$ of the triples. For each triple, the subject fills the subject field while the object field is masked. Models are then fed the prompts and asked to fill in the masked token (i.e., the object). While the templates can affect the predictions of the models, we used the same ones of mLAMA listed in Table \ref{tab:mLAMA_templates} to control for the impact that changing the templates might have on the results. In addition to that, we mapped the templates into questions as shown in Table \ref{tab:GPT3.5_questions} to evaluate the performance of the GPT3.5 model on a subset of DLAMA-v1 (Arab-West).

\begin{table*}[htbp]
    \centering
    \resizebox{\textwidth}{!}{%

    \begin{tabular}{ccc}
\textbf{Cultures} & \textbf{Country} & \textbf{Wikipedia sites used for articles} \\
\midrule
Arab Cultures & 22 countries of the Arab League & Arabic (ar), English (en), French (fr) \\
\midrule
\multirow{20}{*}{Western Cultures} & Australia & English (en) \\
 & Canada & English (en), French (fr) \\
 & New Zealand & English (en), Māori (mi) \\
 & USA & English (en) \\
 \cmidrule{2-3}
 & Andorra & Catalan (ca), English (en) \\
 & Italy & Italian (it), English (en) \\
 & Liechtenstein & German (de), English (en) \\
 & Monaco & French (fr), English (en) \\
 & Portugal & Portuguese (pt), English (en) \\
 & San Marino & Italian (it), English (en) \\
 & Spain & Spanish (es), English (en) \\
 \cmidrule{2-3}
 & Austria & German (de), English (en) \\
 & Belgium & German (de), French (fr), Dutch (nl), English (en) \\
 & France & French (fr), English (en) \\
 & Germany & German (de), English (en) \\
 & Ireland & Irish (ga), English (en) \\
 & Luxembourg & Luxembourgish (lb), French (fr), German (de), English (en) \\
 & Netherlands & Dutch (nl), English (en) \\
 & Switzerland & German (de), French (fr), Italian (it), Romansh (rm), English (en) \\
 & UK & English (en) \\
 \midrule
 \multirow{13}{*}{Asian Cultures} & China & English (en), Chinese (zh)\\
    & Indonesia & English (en),  Indonesian	(id)\\
    & Japan & English (en),  Japanese (ja)\\
    & Malaysia & English (en),  Malay (ms) \\
    & Mongolia & English (en),  Chinese (zh)\\
    & Myanmar & English (en),  Burmese (my)\\
    & North Korea & English (en),  Korean (ko)\\
    & Philippines & English (en)\\
    & Singapore & English (en),  Malay (ms)\\
    & South Korea & English (en),  Korean (ko)\\
    & Taiwan & English (en),  Chinese (zh)\\
    & Thailand & English (en),  Thai (th) \\
    & Vietnam & English (en),  Vietnamese (vi) \\
 \midrule
 \multirow{12}{*}{South American Cultures} & Argentina & English (en), Spanish (es) \\
    & Bolivia & English (en), Spanish (es) \\
    & Brazil & English (en), Portugese (pt) \\
    & Chile & English (en), Spanish (es) \\
    & Colombia & English (en), Spanish (es) \\
    & Ecuador & English (en), Spanish (es) \\
    & Guyana & English (en) \\
    & Paraguay & English (en), Spanish (es) \\
    & Peru & English (en), Spanish (es) \\
    & Suriname & Dutch (nl), English (en) \\
    & Uruguay & English (en), Spanish (es) \\
    & Venezuela & English (en), Spanish (es) \\
\bottomrule
    \end{tabular}%
    }
    \caption{The list of Countries and their respective Wikipedia sites used for representing the four different cultures. The English Wikipedia is used for all the countries.}
    \label{tab:wikipedia_articles_languages}
\end{table*}

\begin{sidewaystable*}[htbp]
    \resizebox{\textwidth}{!}{%
\begin{tabular}{llrll}
\textbf{Predicate} & \textbf{English template} & \textbf{Arabic template} & \textbf{Korean template} & \textbf{Spanish template}\\
\midrule
P17 (Country)&[X] is located in [Y] .& . [Y] \AR{في} [X] \AR{يقع}                                                    &\begin{CJK}{UTF8}{} \CJKfamily{mj} [X]{는} [Y]{에} {있습니다.} \end{CJK} &                        [X] se encuentra en [Y].\\
P19 (Place of birth)&[X] was born in [Y] .& . [Y] \AR{في} [X] \AR{ولد}                                                    &\begin{CJK}{UTF8}{} \CJKfamily{mj} [X]{는} [Y]{에서} {태어났습니다.} \end{CJK} &                               [X] nació en [Y].\\
P20 (Place of death)&[X] died in [Y] .& . [Y] \AR{في} [X] \AR{توفي}                                                   &\begin{CJK}{UTF8}{} \CJKfamily{mj} [X]{는} [Y]{에서} {사망했습니다.} \end{CJK} &                               [X] murió en [Y].\\
P27 (Country of citizenship)&[X] is [Y] citizen .& . [Y] \AR{مواطن} [X]                                                                  &\begin{CJK}{UTF8}{} \CJKfamily{mj} [X]{는} [Y] {시민입니다.} \end{CJK} &                           [X] es [Y] ciudadano.\\
P30 (Continent)&[X] is located in [Y] .& . [Y] \AR{في} [X] \AR{يقع}                                                    &\begin{CJK}{UTF8}{} \CJKfamily{mj} [X]{는} [Y]{에} {있습니다.} \end{CJK} &                        [X] se encuentra en [Y].\\
P36 (Capital)&The capital of [X] is [Y] .& . [Y] \AR{هي} [X] \AR{عاصمة}                                                  &\begin{CJK}{UTF8}{} \CJKfamily{mj} [X]{의} {수도는} [Y]{입니다.} \end{CJK} &                       La capital de [X] es [Y].\\
P37 (Official language)&The official language of [X] is [Y] .& . [Y] \AR{هي} [X] \AR{لـ} \AR{الرسمية} \AR{اللغة}             &\begin{CJK}{UTF8}{} \CJKfamily{mj} [X]{의} {공식} {언어는} [Y]{입니다.} \end{CJK} &                El idioma oficial de [X] es [Y].\\
P47 (Shares border with)&[X] shares border with [Y] .& . [Y] \AR{مع} \AR{الحدود} \AR{في} \AR{تشترك} [X]              &\begin{CJK}{UTF8}{} \CJKfamily{mj} [X]{는} [Y]{와} {(과)} {국경을} {공유합니다.} \end{CJK} &                  [X] comparte frontera con [Y].\\
P103 (Native language)&The native language of [X] is [Y] .& . [Y] \AR{هي} [X] \AR{لـ} \AR{الأصلية} \AR{اللغة}             &\begin{CJK}{UTF8}{} \CJKfamily{mj} [X]{의} {모국어는} [Y]{입니다.} \end{CJK} &                 El idioma nativo de [X] es [Y].\\
P106 (Occupation)&[X] is a [Y] by profession .& \AR{المهنة.} \AR{حسب} [Y] \AR{هي} [X]                                &\begin{CJK}{UTF8}{} \CJKfamily{mj} [X]{는} {직업} {별} [Y]{입니다.} \end{CJK} &                    [X] es una [Y] de profesión.\\
P136 (Genre)&[X] plays [Y] music .& . [Y] \AR{موسيقى} \AR{يعزف} [X]                                               &\begin{CJK}{UTF8}{} \CJKfamily{mj} [X]{는} [Y] {음악을} {재생합니다.} \end{CJK} &                       [X] reproduce música [Y].\\
P190 (Sister city)&[X] and [Y] are twin cities .& \AR{توأمتان.} \AR{مدينتان} [Y] \AR{و} [X]                            &\begin{CJK}{UTF8}{} \CJKfamily{mj} [X]{와} [Y]{는} {쌍둥이} {도시입니다.} \end{CJK} &               [X] e [Y] son ciudades gemelas.\\
P264 (Record label)&[X] is represented by music label [Y] .& . [Y] \AR{الموسيقية} \AR{العلامة} \AR{يمثلها} [X]                     &\begin{CJK}{UTF8}{} \CJKfamily{mj} [X]{는} {음악} {레이블} [Y]{로} {표시됩니다.} \end{CJK} & [X] está representado por el sello musical [Y].\\
P364 (Original language of work)&The original language of [X] is [Y] .& . [Y] \AR{هي} [X] \AR{لـ} \AR{الأصلية} \AR{اللغة}             &\begin{CJK}{UTF8}{} \CJKfamily{mj} [X]{의} {원래} {언어는} [Y]{입니다.} \end{CJK} &               El idioma original de [X] es [Y].\\
P449 (Original network)&[X] was originally aired on [Y] .& . [Y] \AR{على} \AR{الأصل} \AR{في} [X] \AR{بث} \AR{تم} &\begin{CJK}{UTF8}{} \CJKfamily{mj} [X]{는} {원래} [Y]{에} {방영되었습니다.} \end{CJK} &             [X] se emitió originalmente en [Y].\\
P495 (Country of origin)&[X] was created in [Y] .& . [Y] \AR{في} [X] \AR{إنشاء} \AR{تم}                                  &\begin{CJK}{UTF8}{} \CJKfamily{mj} [X]{는} [Y]{에} {작성되었습니다.} \end{CJK} &                             [X] se creó en [Y].\\
P530 (Diplomatic relation)&[X] maintains diplomatic relations with [Y] .& . [Y] \AR{مع} \AR{دبلوماسية} \AR{علاقات} \AR{تقيم} [X]        &\begin{CJK}{UTF8}{} \CJKfamily{mj} [X]{는} [Y]{와의} {외교} {관계를} {유지합니다.} \end{CJK} &   [X] mantiene relaciones diplomáticas con [Y].\\
P1303 (Instrument)&[X] plays [Y] .& . [Y] \AR{يلعب} [X]                                                                   &\begin{CJK}{UTF8}{} \CJKfamily{mj} [X]{는} [Y]{를} {재생합니다.} \end{CJK} &                              [X] reproduce [Y].\\
P1376 (Capital of)&[X] is the capital of [Y] .& . [Y] \AR{عاصمة} \AR{هي} [X]                                                  &\begin{CJK}{UTF8}{} \CJKfamily{mj} [X]{는} [Y]{의} {수도입니다.} \end{CJK} &                       [X] es la capital de [Y].\\
P1412 (Languages spoken or published)&[X] used to communicate in [Y] .& . [Y] \AR{في} \AR{للتواصل} [X] \AR{يستخدم}                            &\begin{CJK}{UTF8}{} \CJKfamily{mj} [X]{는} [Y]{에서} {통신하는} {데} {사용됩니다.} \end{CJK} &                   [X] solía comunicarse en [Y].\\
\bottomrule
\end{tabular}%
    }
\caption{mLAMA's templates that are also adapted in DLAMA.}
\label{tab:mLAMA_templates}
\end{sidewaystable*}

\begin{sidewaystable*}[htbp]
    \resizebox{\textwidth}{!}{%
\begin{tabular}{llr}
\textbf{Predicate} & \textbf{English Question} & \textbf{Arabic Question}\\
\midrule
P30 (Continent) &                                    Where is "[X]" located in? Reply with a name of a continent only. &\AR{فقط} \AR{قارة} \AR{باسم} \AR{أجب} \AR{؟} "]X[" \AR{يقع} \AR{أين}                                                                                                                  \\
P36 (Capital) &                                  What is the capital of "[X]"? Reply with the name of the city only. &\AR{فقط} \AR{المدينة} \AR{باسم} \AR{أجب} \AR{؟} "]X[" \AR{عاصمة} \AR{هي} \AR{ما}                                                                                              \\
P37 (Official Language) &                           What is the official language of "[X]"? Reply with the language name only. &\AR{فقط} \AR{لغة} \AR{باسم} \AR{أجب} \AR{؟} "]X[" \AR{ل} \AR{الرسمية} \AR{اللغة} \AR{هي} \AR{ما}                                                              \\
P47 (Shares border with) &                   What is the country that shares border with "[X]"? Reply with a country name only. &\AR{فقط} \AR{دولة} \AR{باسم} \AR{أجب} \AR{؟} "]X[" \AR{مع} \AR{حدودها} \AR{تشترك} \AR{التي} \AR{الدولة} \AR{هي} \AR{ما}                       \\
P190 (Sister city) &                                What is the twin city of "[X]"? Reply with the name of the city only. &\AR{فقط} \AR{المدينة} \AR{باسم} \AR{أجب} \AR{؟} "]X[" \AR{لمدينة} \AR{التوأم} \AR{المدينة} \AR{هي} \AR{ما}                                                    \\
P530 (Diplomatic realtion) & What is the country that maintains dimplomatic relations with "[X]"? Reply with a country name only. &\AR{فقط} \AR{دولة} \AR{باسم} \AR{أجب} \AR{؟} "]X[" \AR{مع} \AR{دبلوماسية} \AR{علاقات} \AR{تقيم} \AR{التي} \AR{الدولة} \AR{هي} \AR{ما} \\
P1376 (Capital of) &                   What is the country of which the capital is "[X]"? Reply with a country name only. &\AR{فقط} \AR{دولة} \AR{باسم} \AR{أجب} \AR{؟} "]X[" \AR{عاصمتها} \AR{التي} \AR{الدولة} \AR{هي} \AR{ما}                                                         \\
\bottomrule
\end{tabular}    %
    }
\caption{The mapping of six of mLAMA's templates to questions that can be used to evaluate the GPT3.5-turbo model.}
\label{tab:GPT3.5_questions}
\end{sidewaystable*}




\end{document}